\documentclass[10pt,twocolumn,letterpaper]{article}
\usepackage[pagenumbers]{cvpr}
\usepackage{amsmath,amssymb,booktabs,graphicx,array}
\usepackage{tikz}

\usepackage{colortbl}
\definecolor{progressbest}{RGB}{255,235,215}

\usepackage{tabularx}

\DeclareRobustCommand{\projname}{\textup{\textsc{Virgo}}}
\newcommand{\papertitle}{\projname: A \textbf{Vi}suomoto\textbf{r} \textbf{G}eneralist for Whole-Body Human\textbf{o}id Interaction}

\definecolor{cvprblue}{rgb}{0.21,0.49,0.74}
\usepackage[pagebackref,breaklinks,colorlinks,allcolors=cvprblue]{hyperref}
\def\confName{CVPR}\def\confYear{2027}
\title{\papertitle}
\author{
Jinxian Zhou$^{1}$\qquad Lin Wu$^{1,2}$\qquad Xiangyu Huang$^{1}$\qquad Yifan Sun$^{1}$\\
Siyuan Li$^{1}$\qquad Peiqi Yu$^{1}$\qquad Ziyu Xu$^{1}$\qquad Changliu Liu$^{1}$\\
$^{1}$Carnegie Mellon University\qquad $^{2}$University of Glasgow
}
\newcommand{\teaserfigure}{
\begin{minipage}{\textwidth}
\centering
\includegraphics[width=0.98\textwidth]{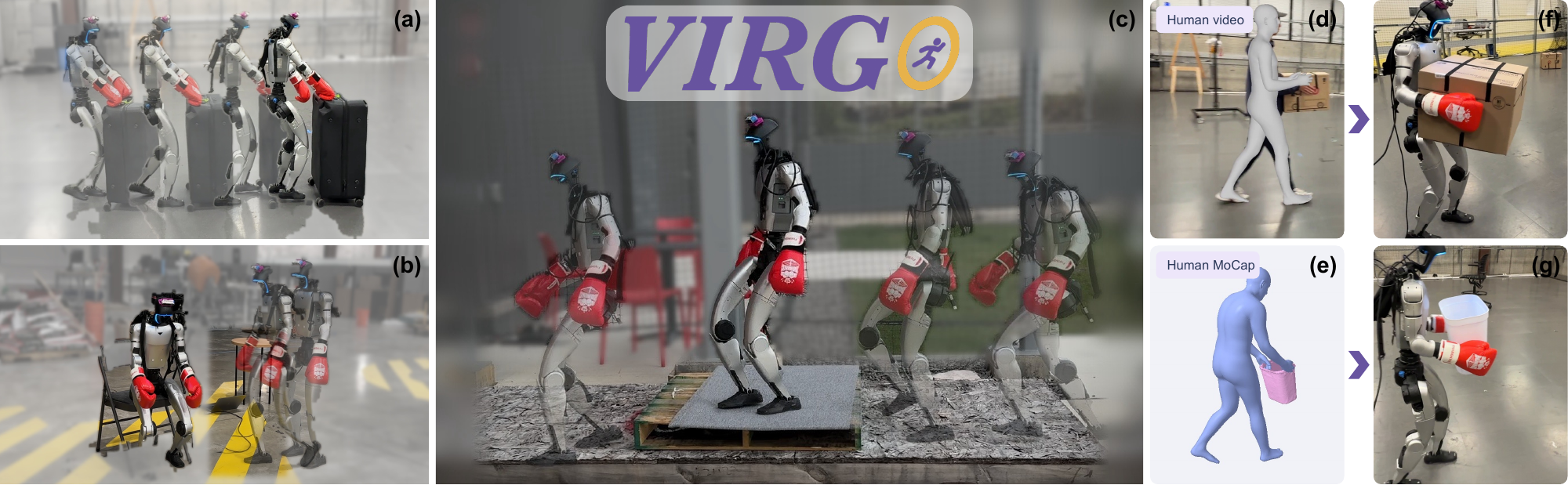}
\captionsetup{hypcap=false}
\captionof{figure}{\textbf{\projname{}} enables a single visuomotor policy to perform diverse whole-body interactions, transferring zero-shot from simulation: (a) object interaction, (b) sitting, and (c) climbing, shown as time-lapses. The policy follows references from either (d) a monocular human video or (e) human MoCap, with the corresponding robot executions shown in (f) and (g).}

\label{fig:teaser}
\end{minipage}\par\vspace{12pt}
}
\newcommand{\pipelinefigure}{
\begin{figure*}[t]
\centering
\includegraphics[width=\textwidth]{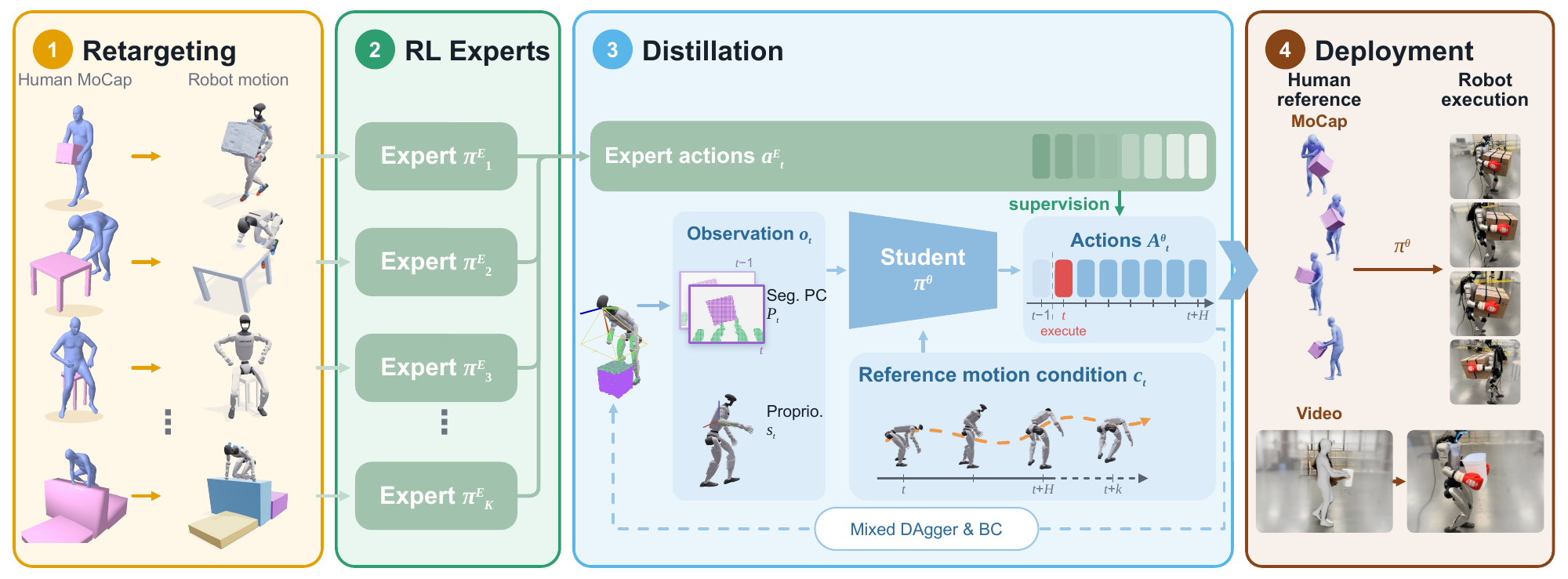}
\caption{\textbf{\projname{} overview.} 1) Human demonstrations are retargeted into robot references. 2) The references are tracked by state-based RL experts. 3) The experts' rollouts are then distilled into one reference-conditioned student policy that observes a segmented point cloud and proprioception, via mixed DAgger\&BC. 4) The generalist student then follows new MoCap or video references from egocentric perception.}
\label{fig:pipeline}
\end{figure*}
}
\newcommand{\paperabstract}{
\begin{abstract}
A key open challenge in humanoid robotics is learning a single policy that autonomously performs diverse whole-body interactions using onboard perception.
Existing motion trackers often lack visual feedback, whereas perceptive policies typically specialize in individual task families.
Contact-rich robot data remains costly to collect.
We introduce \projname{}, a visual sim-to-real framework that learns diverse interactions from human demonstrations without teleoperated robot training data.
Our approach trains state-based Reinforcement Learning (RL) experts on retargeted motion-capture (MoCap) trajectories and distills them into one reference-conditioned visuomotor diffusion policy through mixed Behavior Cloning (BC) and online DAgger.
The resulting policy follows MoCap or video-derived references, using egocentric visual feedback to maintain the intended interaction in a closed loop.
Experiments span moving-object interaction, climbing, and sitting, with a mean trajectory completion ratio of 0.828 on held-out object-interaction references, outperforming the evaluated deployable baselines.
The same frozen policy transfers zero-shot to a Unitree G1 across human references and scene variations, without task-specific fine-tuning.
Project website: \url{https://virgo-humanoid.github.io/}.
\end{abstract}
}
\makeatletter
\apptocmd{\@maketitle}{\teaserfigure}{}{\PackageError{virgo}{Cannot attach teaser to title}{}}
\makeatother
\begin{document}
\maketitle
% Queue the overview on page 1 so the double-column float tops page 2.
% \pipelinefigure
\paperabstract
% =====================================================================
\section{Introduction}
\label{sec:intro}
% P1 -- Goal 
Learning a single humanoid policy for diverse whole-body interactions requires both versatile motion control and feedback about the surrounding scene.
Carrying an object, climbing onto a platform, and sitting on a chair involve different contact patterns, for which accurate body-motion tracking alone is insufficient.
A robot may continue the prescribed movement even after losing an object or missing a supporting surface.
The challenge is to cover these varied behaviors within one policy while adapting execution to the current interaction geometry.

% Humanoid robots are often viewed as a natural embodiment for general-purpose physical intelligence, given their human-like morphology and versatility.
% Realizing this promise requires more than walking or replaying motions in free space: the robot must use its whole body to accomplish some meaningful and useful behaviors with interactions, like carrying a box, stepping onto a platform or sitting.
% P2 -- what is missing in existing works 
Existing approaches address parts of this challenge.
Motion trackers reproduce diverse body motions but typically lack visual feedback about their surroundings~\cite{chen2025gmt,liao2025beyondmimic,luo2025sonic}.
Teleoperation systems allow human operators to guide execution~\cite{he2024h2o,he2024omnih2o,ze2025twist}, but collecting contact-rich robot demonstrations requires sustained operator effort.
Perceptive policies respond to scene geometry, yet commonly specialize in individual task families such as parkour or loco-manipulation~\cite{zhuang2024parkourlearning,wu2026php,he2025viral}.
A core missing piece is autonomous, generalist humanoid control that combines broad interaction coverage with onboard visual feedback.

% A core missing piece is autonomous and generalist humanoid control, that performs diverse meaningful behaviors from onboard perception. 
% Most existing systems either track motions without perceiving the scene~\cite{chen2025gmt,liao2025beyondmimic,luo2025sonic}, rely on a human operator in the control loop~\cite{he2024h2o,he2024omnih2o,fu2025humanplus,ze2025twist}, 
% or learn a separate perceptive policy for each skill family~\cite{zhuang2024parkourlearning,wu2026php,he2025viral}.

% P3 -- Human data is abundant but not executable: mocap --> executable actions via RL --> teacher-student visual sim-to-real
Human demonstrations offer a source of diverse interaction behaviors without requiring teleoperated robot training data.
Motion-capture (MoCap) datasets and videos record how people manipulate objects and use environmental supports~\cite{mahmood2019amass,li2023omomo,taheri2020grab,bhatnagar2022behave}.
However, differences in morphology and dynamics prevent these motions from being directly executed by a humanoid~\cite{araujo2025gmr,yang2025omniretarget}.
Simulation provides a way to bridge this gap: retargeting maps human demonstrations to robot references, and Reinforcement Learning (RL) converts these references into dynamically feasible behaviors~\cite{peng2018deepmimic,luo2023phc,weng2025hdmi}.
Teacher--student distillation then transfers privileged-state supervision to policies using deployable observations~\cite{rusu2015distillation,pinto2017asymmetric,he2025viral}.
Extending this approach across interaction families requires consolidating diverse experts into one policy while specifying which behavior to execute.

We introduce A \textbf{VI}suomoto\textbf{R} \textbf{G}eneralist for Whole-Body Human\textbf{O}id Interaction, \textbf{\projname{}}, 
% \changliu{update according to the new title}
a visual sim-to-real framework that learns one reference-conditioned humanoid policy through \emph{visual expert-to-generalist distillation} (Fig.~\ref{fig:pipeline}).
State-based RL experts first learn groups of retargeted MoCap trajectories, then supervise a shared visuomotor diffusion policy~\cite{chi2023diffusion,ze2024dp3}.
Mixed Behavior Cloning (BC) and online DAgger~\cite{ross2011dagger,he2025viral} combine nominal expert demonstrations with supervision at student-visited states.
The motion reference specifies the intended behavior, while proprioception and egocentric point clouds provide feedback for its execution.
The same frozen policy follows new MoCap or video-derived references without fine-tuning or human corrections.
% We present \textbf{\projname{}}, a framework for learning a generalist humanoid policy from diverse human demonstrations through visual expert-to-generalist distillation (Fig.~\ref{fig:pipeline}).
% During RL training, \projname{} uses motion-capture dataset as reference trajectories in simulation.
% For each group of compatible trajectories, \projname{} trains a trajectory-conditioned RL expert with a shared reward, which converts the demonstrated whole-body interaction into physically feasible robot behavior while compensating for retargeting errors and contact dynamics.

% The state-based experts transform mocap references into robot execution data, which we distill into a vision-based diffusion policy~\cite{chi2023diffusion,ze2024dp3}.
% The distilled student policy closes the loop from robot proprioception and segmented egocentric observations while following the motion reference: dense, timestep-aligned reference features guide local action generation, and sparse look-ahead features anticipate upcoming changes in contact and support.
% To cover states induced by student errors, we combine Behavior Cloning (BC) environments with mixed online DAgger~\cite{ross2011dagger}, where the probability of executing expert actions decays exponentially over iterations~\cite{he2025viral}.
% We find this mixture more effective than either pure BC or pure DAgger.

% P5 -- Evaluation and results
We evaluate moving-object interaction, climbing, and sitting using references from OMOMO~\cite{li2023omomo}, OmniRetarget~\cite{yang2025omniretarget}, and AMASS~\cite{mahmood2019amass}.
On held-out object-interaction references, \projname{} achieves a mean trajectory completion ratio of 0.828, outperforming the evaluated deployable baselines.
A qualitative recovery comparison shows that visual feedback helps preserve the intended interaction after disturbances, while hardware experiments demonstrate zero-shot transfer to a Unitree G1 across human references and scene variations. 
Our contributions are:
\begin{itemize}
    \item \textbf{Expert-to-generalist learning.}
    We consolidate motion-group RL experts into one visuomotor
    policy, learning diverse interactions from human demonstrations
    without teleoperated robot training data.

    \item \textbf{Visual closed-loop interaction.}
    Robot-relative references specify the intended behavior,
    while egocentric 3D observations provide feedback to
    maintain the interaction across task families.

    \item \textbf{Generalization and zero-shot sim-to-real.}
    Our policy surpasses retargeting and a state-of-the-art
    motion tracker in held-out object-interaction Progress
    in simulation.
    The same frozen policy executes unseen MoCap and
    video-derived references zero-shot on a physical humanoid.
\end{itemize}

\pipelinefigure
\section{Related Work}
\label{sec:related}
\paragraph{Human motion to humanoid control.}
Human motion provides supervision for physics-based tracking~\cite{peng2018deepmimic,luo2023phc} and reusable motion priors~\cite{peng2021amp,peng2022ase}.
Transferring this supervision across embodiments requires retargeting: GMR addresses human--robot morphology differences~\cite{araujo2025gmr}, while OmniRetarget preserves body--object--terrain relations~\cite{yang2025omniretarget}.
HDMI further reconstructs human--object references from videos for interactive whole-body control~\cite{weng2025hdmi}.
Another route uses human motion for teleoperation and subsequent policy learning~\cite{fu2025humanplus,he2024omnih2o,ze2025twist}.
Building on interaction-aware retargeting, \projname{} uses human demonstrations to train motion-group experts entirely in simulation, then consolidates them into a shared visuomotor policy without teleoperated robot training data.

% OmniRetarget preserves body--object--terrain relations~\cite{yang2025omniretarget}, and HDMI reconstructs human--object references from video~\cite{weng2025hdmi}. 
% HumanPlus, OmniH2O, and TWIST connect human motion to humanoid teleoperation and learning~\cite{fu2025humanplus,he2024omnih2o,ze2025twist}. 
% We consolidate retargeted motion-group experts into a shared visuomotor controller.
% \vspace{-3mm}
\paragraph{Generalist policy learning.}
Generalist humanoid controllers broaden motion coverage through expressive motion learning~\cite{cheng2024exbody,ji2024exbody2}, unified control interfaces~\cite{he2024hover,tessler2024maskedmimic}, and large-scale motion tracking~\cite{chen2025gmt,liao2025beyondmimic,luo2025sonic}.
Recent work further improves reference feasibility and learning efficiency~\cite{li2026omnitrack,liao2026supermimic}, while BFM learns reusable behaviors through large-scale pretraining~\cite{zeng2025bfm}.
However, broad motion coverage alone does not ensure that a controller maintains contact with moving objects or environmental supports.
\projname{} targets this interaction-level generality: one reference-conditioned policy combines diverse motion guidance with egocentric visual feedback to maintain the intended interaction across task families.

\paragraph{Perceptive sim-to-real control.}
Sim-to-real RL enables robust locomotion through domain randomization and adaptation~\cite{hwangbo2019anymal,kumar2021rma,rudin2022walkinminutes,tobin2017randomization,peng2018dynamics,he2025asap}.
To account for surrounding geometry, perceptive controllers incorporate depth or terrain observations, often through privileged teacher--student learning~\cite{rusu2015distillation,pinto2017asymmetric,miki2022perceptive,cheng2024extremeparkour,zhuang2024parkourlearning,zhuang2026parkour,wu2026php}.
VideoMimic connects reconstructed human motions and scenes to contextual humanoid control~\cite{allshire2025videomimic}, while VIRAL scales visual distillation for loco-manipulation using mixed BC and DAgger~\cite{he2025viral,ross2011dagger}.
We build on this visual distillation paradigm, but consolidate multiple motion-group experts through a common reference interface.
The resulting student shares its parameters across moving-object interaction, climbing, and sitting, with visual feedback closing the execution loop for the specified behavior.

% \paragraph{Sim-to-real for whole-body control.}
% Sim-to-real reinforcement learning has enabled robust blind locomotion on legged robots and humanoids~\cite{hwangbo2019anymal,kumar2021rma,rudin2022walkinminutes}, with domain randomization~\cite{tobin2017randomization,peng2018dynamics} and dynamics alignment~\cite{he2025asap} narrowing the reality gap.
% Proprioceptive policies, however, lack awareness of their surroundings.
% Perceptive methods add depth or elevation-map observations, typically by distilling a privileged teacher into a vision-based student~\cite{rusu2015distillation,pinto2017asymmetric,miki2022perceptive,cheng2024extremeparkour,zhuang2024parkourlearning,zhuang2026parkour,wu2026php}.
% VideoMimic learns contextual humanoid skills from reconstructed human motions and scenes~\cite{allshire2025videomimic}, and VIRAL scales visual teacher--student distillation with mixed DAgger and BC to humanoid loco-manipulation~\cite{he2025viral,ross2011dagger}.
% These methods typically train one perceptive policy per task family.
% \projname{} extends visual sim-to-real to multi-task whole-body interaction by distilling many motion-group experts into a single reference-conditioned student.

% \changliu{and why is it important for generalists? Also need to cite BFM here.}

\section{Method}
\label{sec:method}

\projname{} learns a shared visuomotor controller from diverse human demonstrations (Fig.~\ref{fig:pipeline}). 
Retargeted motions first supervise privileged RL experts. Their actions then train a reference-conditioned diffusion student through a mixture of BC and online DAgger. 
At deployment, the student follows a supplied motion reference using visual and proprioceptive feedback.
\vspace{-2mm}
\paragraph{Problem formulation.}
Let $\mathcal R$ be a retargeted trajectory containing robot poses, joint states, and available scene annotations.
At time $t$, the student receives observation history $O_t=\{o_{t-h+1},\ldots,o_t\}$, where $o_t=(v_t,s_t)$ comprises egocentric geometry and proprioception, and a robot-motion reference window $C_t=\Phi(\mathcal R,t;h,H)$.
Here, $\Phi$ extracts reference states spanning the observation history of length $h$ and the look-ahead horizon $H$.
The student predicts an action window
\begin{equation}
 A_t\sim\pi_\theta(\,\cdot\mid O_t,C_t),
\label{eq:policy}
\end{equation}
where $\theta$ denotes the student policy parameters.
The full reference is prepared before deployment, but the policy receives only the local window $C_t$ of relative torso poses and joint states.
It executes the current aligned action and adapts subsequent
actions using the latest onboard observations, without a human operator's corrections.
The reference specifies the intended behavior, while sensory feedback closes the execution loop.
The student parameters are shared across tasks.
% Let $\mathcal R$ be a retargeted trajectory containing robot poses, joint states, and available scene annotations. 
% At time $t$, the student receives observation history $O_t=\{o_{t-h+1},\ldots,o_t\}$, where $o_t=(v_t,s_t)$ comprises egocentric geometry and proprioception, and a robot-motion reference window $C_t=\Phi(\mathcal R,t)$. It predicts an action window
% \begin{equation}
%  A_t\sim\pi_\theta(\,\cdot\mid O_t,C_t).
% \label{eq:policy}
% \end{equation}
% Here, $h$ is the observation-history length, $H$ is the reference look-ahead horizon, \changliu{should $H$ be a parameter for $C_t$?} 
% Re: it's constant
% $\Phi$ extracts the reference window, and $\theta$ denotes the student policy parameters.
% The full reference is prepared before deployment, but the policy receives only a local window $C_t$ of relative torso poses and joint states spanning the observation history and the reference horizon $H$.

\subsection{RL Expert Training from MoCap Data}
\label{sec:experts}
\paragraph{Importing MoCap data into simulation.}
We import human motion and the associated scene geometry into simulation, then retarget the demonstrations to the humanoid using OmniRetarget~\cite{yang2025omniretarget}. 
Its interaction mesh couples human body keypoints with sampled
object and terrain points, and minimizing mesh deformation
under robot kinematic constraints preserves task-relevant
spatial relations across embodiments.
In particular, retaining hand--object and body--environment contacts is essential for object interaction, climbing, and sitting, where matching body poses alone can break the demonstrated interaction. 
The resulting kinematic references contain robot joint and torso trajectories, together with available object trajectories and contact annotations. 
% We use 232 sequences from OMOMO~\cite{li2023omomo}, 10 sequences from AMASS~\cite{mahmood2019amass}, and 20 climbing sequences from OmniRetarget~\cite{yang2025omniretarget}.
We use 232 original reference trajectories from
OMOMO~\cite{li2023omomo}, 10 sitting trajectories from
AMASS~\cite{mahmood2019amass}, and 20 climbing trajectories
from OmniRetarget~\cite{yang2025omniretarget} for training.
We apply task-specific trajectory randomization and augmentation during RL expert training and student distillation; sitting collection retains fixed motion--stool pairs without trajectory randomization (see supplementary material).
Held-out evaluation trajectories are excluded from both stages.
\vspace{-3mm}
% \begin{figure}[t]
% \centering
% \includegraphics[width=\linewidth,pagebox=cropbox]{figures/RL.pdf}
% \caption{\textbf{Reference-state initialization.} Episodes start from configurations sampled along the reference, exposing experts to all motion phases.}
% \label{fig:experts}
% \end{figure}
% \vspace*{-3mm}
\paragraph{Training RL experts.}
For trajectory group $k$, the expert policy $\pi_k^E$ observes the privileged simulator state $x_t^{\mathrm{priv}}$ (robot and available scene states) and its desired-motion command $C_t^E$. At time $t$, it outputs a 29-dimensional joint action $a_t^E\sim\pi_k^E(\cdot\mid x_t^{\mathrm{priv}},C_t^E)$, converted to PD targets as
\begin{equation}
 q_t^{\mathrm{des}}=q^{\mathrm{default}}+Sa_t^E,
\label{eq:expert_action}
\end{equation}
where $q^{\mathrm{default}}$ is the nominal joint configuration and $S$ contains joint-specific action scales. The policy learns feedback corrections for deviations from the reference while maintaining whole-body balance. Experts and student share this action interface, allowing their behaviors to be distilled into a common output space.

Across multiple MoCap trajectories covering different tasks, we use the same reward components, activating object tracking only when a moving object is present:
\begin{equation}
\begin{aligned}
 r_t={}&r_t^{\mathrm{body}}+m_o\lambda_o r_t^{\mathrm{object}}
       +\lambda_c r_t^{\mathrm{contact}}-r_t^{\mathrm{reg}},\\
 m_o={}&\begin{cases}
 1, & \text{object interaction},\\
 0, & \text{climbing or sitting}.
 \end{cases}
\end{aligned}
\label{eq:reward}
\end{equation}

% $\lambda_o$ and $\lambda_c$ weight object tracking
% and contact rewards, respectively.
% Specifically, $r_t^{\mathrm{body}}$ tracks reference torso
% and body poses and velocities using exponential rewards
% $\exp(-d^2/\sigma^2)$, where $d$ is the corresponding
% tracking deviation and $\sigma>0$ its tolerance scale.
% The object reward $r_t^{\mathrm{object}}$ tracks the
% moving-object pose when $m_o=1$, while
% $r_t^{\mathrm{contact}}$ encourages annotated robot--object
% or robot--environment contacts.
% The regularization term $r_t^{\mathrm{reg}}$ penalizes
% abrupt actions, joint-limit violations, and unintended
% contacts.
% The contact term is omitted when annotations are unavailable.
$\lambda_o$ and $\lambda_c$ weight the object and contact rewards, respectively.
Specifically, $r_t^{\mathrm{body}}$ tracks the reference torso and body pose and velocity, using exponential rewards $\exp(-d^2/\sigma^2)$, where $d$ is the tracking deviation and $\sigma>0$ its corresponding tolerance scale; $r_t^{\mathrm{object}}$ tracks the demonstrated moving-object pose when $m_o=1$.
The contact reward $r_t^{\mathrm{contact}}$ favors annotated robot--object or robot--environment contacts, while $r_t^{\mathrm{reg}}$ discourages abrupt actions, joint-limit violations, and unintended contacts.
The contact term is omitted when annotations are unavailable.

We train 16 motion-group experts, grouping references by
object category and task for object interaction, stool height
for sitting, and terrain configuration for climbing.
Each expert tracks multiple reference sequences within its
group, conditioned on the sampled reference.
Experts are optimized with PPO~\cite{schulman2017ppo} in Isaac Sim at 50\,Hz, using 4096 parallel environments for object interaction and 1024 for climbing and sitting. 
Object-interaction and climbing experts use adaptive reference-phase initialization, while sitting experts start at the first frame. 
The trained experts subsequently provide action supervision for the shared student. 

Under the evaluated training budget, SuperMimic~\cite{liao2026supermimic} reaches only moderate success on our object-interaction set after 686M transitions (72 GPU-hours), motivating grouped expert training followed by distillation (see supplementary material).
% We find that directly training one generalist tracker on the full interaction set is less efficient: a world-model-based generalist (SuperMimic~\cite{liao2026supermimic}) plateaus at a moderate success rate on OMOMO objects despite substantially more training (686M transitions, 72 GPU-hours), and long-tail objects such as chairs and suitcases remain poorly learned (see the supplementary material for per-object results).

\begin{figure}[t]
\centering
\includegraphics[width=0.98\linewidth]{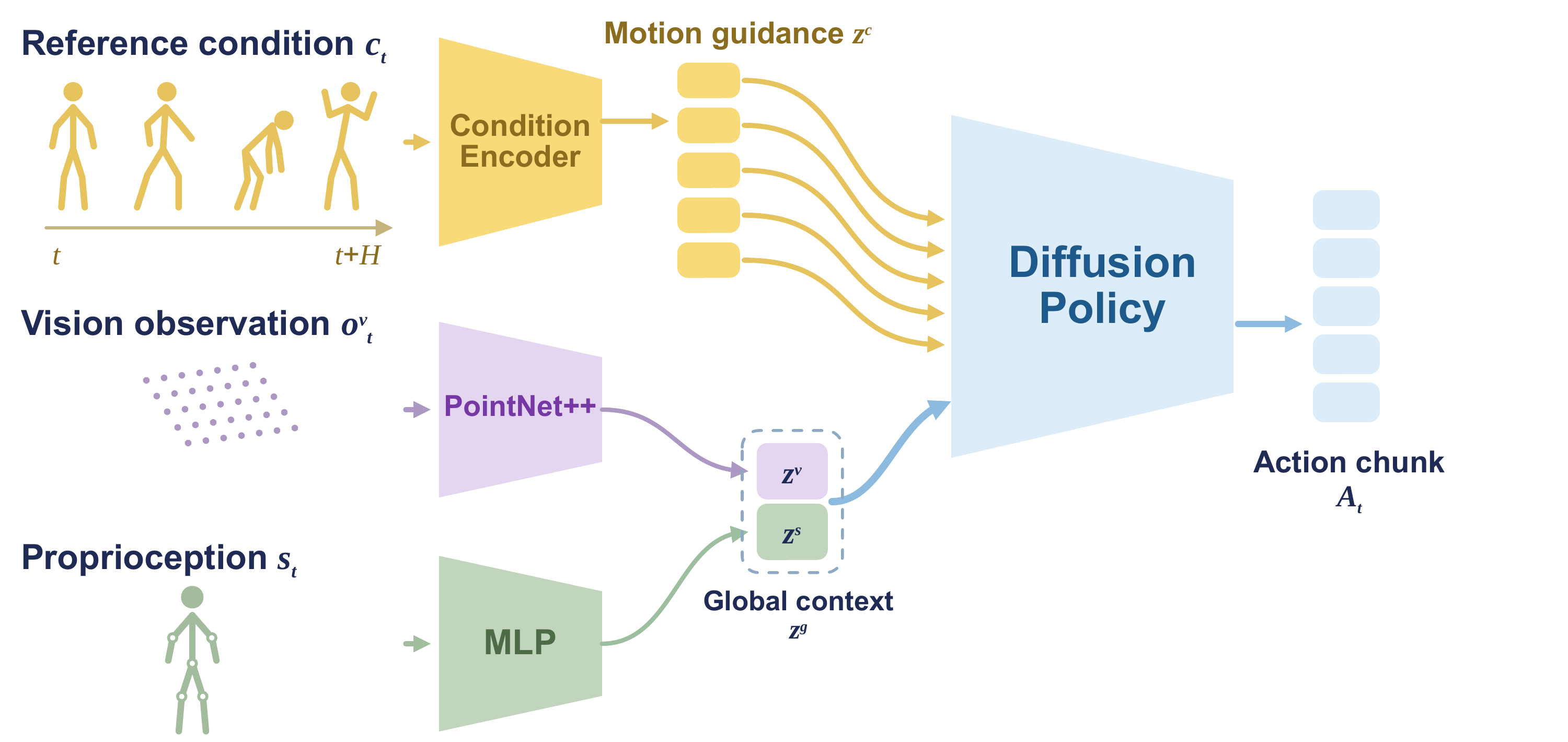}
\caption{\textbf{Reference-conditioned diffusion student.} Reference motion features $z^c$ guide the action window, while point-cloud and proprioceptive features form the global context $z^g$.}
\label{fig:architecture}
\end{figure}
\subsection{Reference-Conditioned Policy Distillation}
\label{sec:distill}
The experts provide dynamically feasible actions but they are tied to individual task groups and depend on privileged simulation states. We therefore set them as teachers, and distill their diverse reference-following behaviors into one deployable student policy that uses visual observations and a common robot-motion condition. Rollout collection retains randomizations.
\vspace{-3mm}
% \changliu{what is this short name for?}
\paragraph{Policy architecture.}
The student combines a segmented point cloud, proprioception, and a window of desired robot motion in a conditional diffusion policy~\cite{ho2020ddpm,chi2023diffusion} (Fig.~\ref{fig:architecture}). We instantiate it as a DP3-style policy~\cite{ze2024dp3} with a PointNet++ point-cloud encoder~\cite{qi2017pointnetpp}, which maps the point cloud to visual features $z^v$, and an MLP that maps proprioception to $z^s$. A condition encoder maps the reference window, including its sparse look-ahead frames, to timestep-aligned motion features $z^c$, while $z^v$ and $z^s$ form the global context $z^g$. A conditional temporal U-Net~\cite{chi2023diffusion,ze2024dp3} uses $z^c$ and $z^g$ to denoise a joint-action window. Dense guidance aligns each predicted action with the near-term reference, while look-ahead anticipates upcoming changes. Only the current aligned action is executed before the policy replans from new observations.

The observation stream contains a short history of proprioception $s_t$ (joint positions and velocities $q_t,\dot q_t$, previous actions $a_{t-1}$, torso angular velocity $\omega_t$, and projected gravity $g_t$), together with an egocentric segmented point cloud $o^v_t$. 
The humanoid's onboard camera captures RGB-D images, which are segmented into robot and scene geometry, and then back-projected into 3D (Fig.~\ref{fig:pipeline}). 
Point-set representations provide a direct interface to 3D geometry~\cite{qi2017pointnetpp,ze2024dp3}, representing nearby objects, robot body parts, and supporting surfaces.
\vspace{-3mm}
\begin{figure}[t]
\centering
\includegraphics[width=0.95\linewidth]{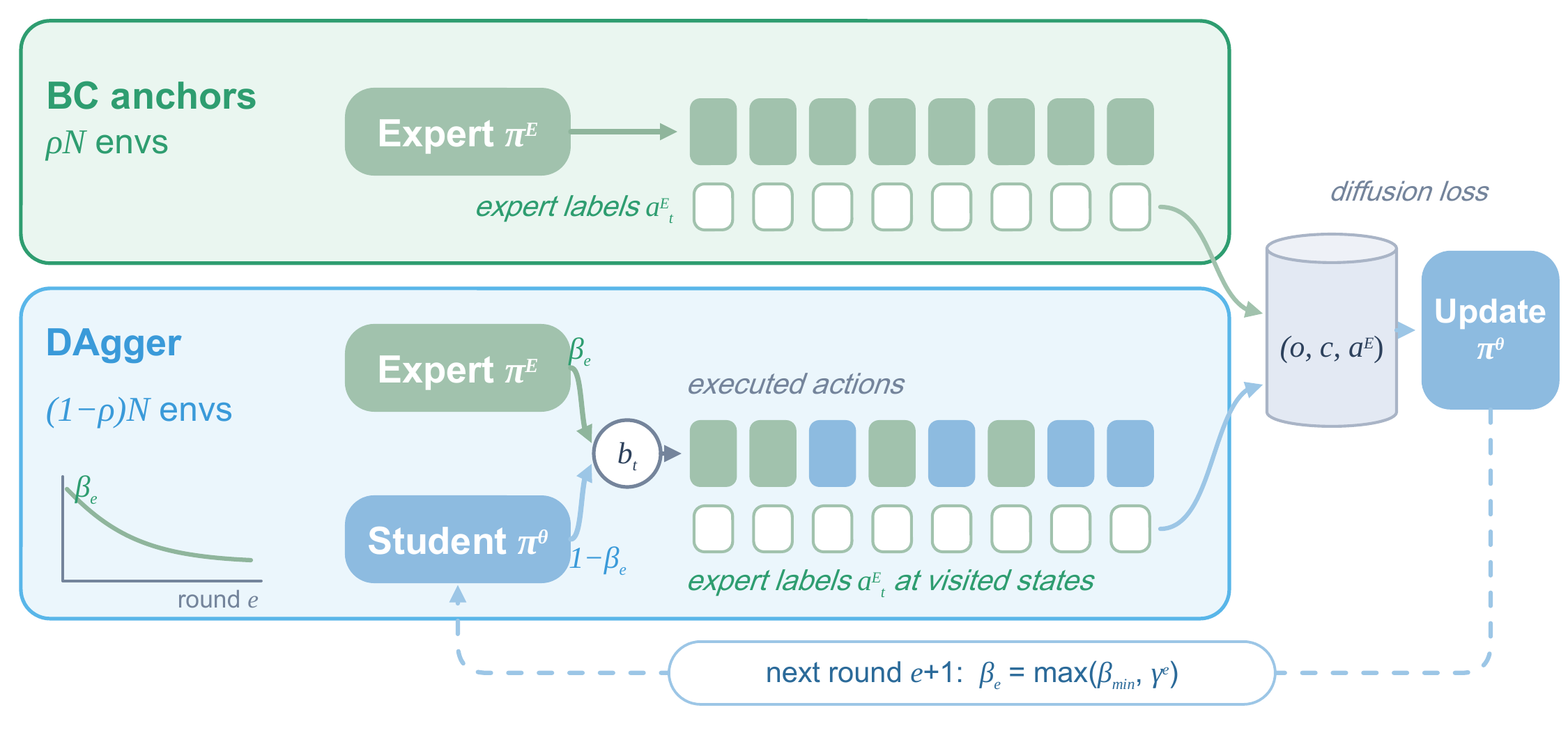}
\caption{\textbf{Mixed DAgger\&BC distillation.} BC environments always execute expert actions, and DAgger environments execute the expert with a decaying probability $\beta_e$. Both provide expert labels at visited states.}
\label{fig:collection}
\vspace{-3mm}
\end{figure}
\begin{figure*}[t]
\centering
\includegraphics[width=\textwidth]{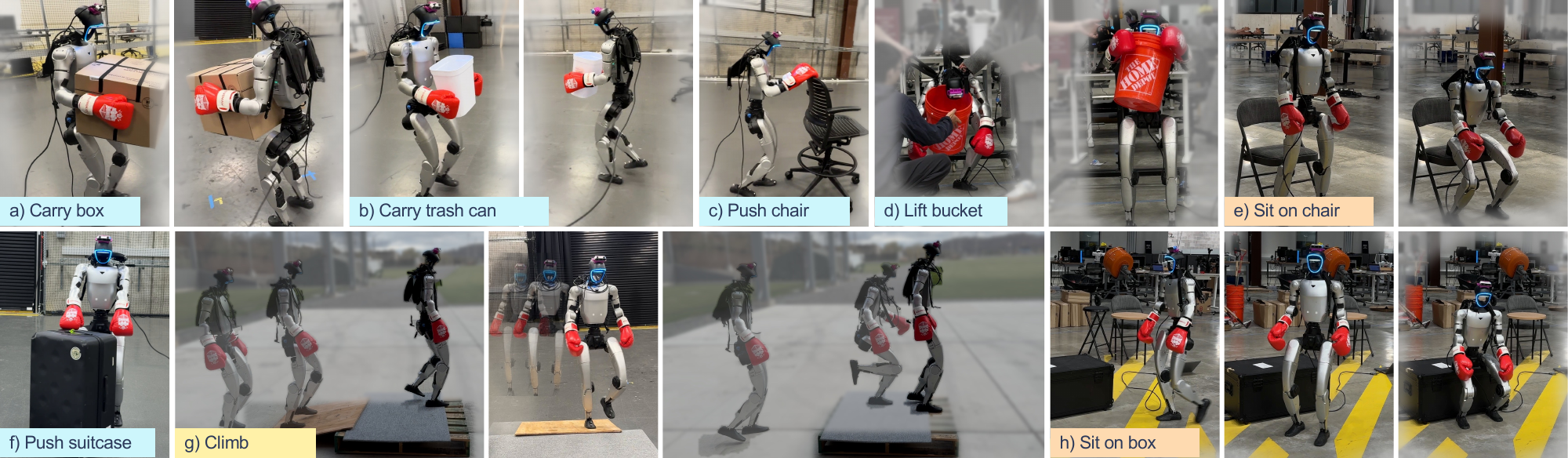}
\caption{\textbf{Multi-task hardware execution with a shared student.} One frozen \projname{} checkpoint on a Unitree G1 (a) carries a box, (b) carries a trash can, (c) pushes a chair, (d) bends down to grasp a bucket and lifts it, (e) sits on a chair, (f) pushes a suitcase, (g) climbs a ramp, a platform shown in front view, and a step,
with the ramp and step panels shown as time-lapse composites, and (h) sits on a box. Other frames are sampled in temporal order from left to right. The same policy weights are used in every panel. }
\label{fig:real}
\end{figure*}
% \vspace{-3mm}
\paragraph{DAgger\&BC Mixture.}
Expert-only rollouts capture nominal tracking but miss states induced by student errors. Following the motivation of DAgger\&BC mixtures~\cite{ross2011dagger,he2025viral}, we reserve a fraction $\rho$ of environments as BC environments that always execute expert actions. In the remaining DAgger environments, an expert or the current student controls each step (Fig.~\ref{fig:collection}):
\begin{equation}
\begin{gathered}
 a_t^{\mathrm{exec}}=
 \begin{cases}
 a_t^E, & \text{BC env.\ or }b_t=1,\\
 a_t^S, & \text{otherwise},
 \end{cases}\\
 b_t\sim\operatorname{Bernoulli}(\beta_e),\qquad
 \beta_e=\max(\beta_{\min},\gamma^e),
\end{gathered}
\label{eq:mix}
\end{equation}
% where $a_t^S$ is the student's current action, $e$ indexes
% collection rounds, $\gamma\in(0,1)$ controls the decay,
% and $\beta_{\min}$ is the minimum expert probability.
% At every visited state, the matched expert supplies
% $a_t^E$ as supervision, regardless of which policy acts.
% For a normalized expert action window $A^E$, we sample
% $\epsilon\sim\mathcal{N}(0,I)$ and a diffusion timestep
% $\ell$ uniformly from the training schedule, and form
% $A^\ell=\sqrt{\bar{\alpha}_\ell}A^E+
% \sqrt{1-\bar{\alpha}_\ell}\epsilon$,
% where $\bar{\alpha}_\ell$ is the cumulative
% signal-retention coefficient.
% The denoising network $f_\theta$ predicts the clean
% action window through
% \begin{equation}
%  \mathcal L(\theta)=\mathbb E\!\left[
%  \frac{\|M\odot(f_\theta(A^\ell,\ell,O,C)-A^E)\|_2^2}
%  {\max(1,\sum_j M_j)}
%  \right].
% \label{eq:diffusion}
% \end{equation}

where $a_t^S$ is the student's current action, $e$ indexes collection rounds, $\gamma\in(0,1)$ controls the decay, and $\beta_{\min}$ is the minimum expert probability.
Both branches provide expert labels at visited states, regardless of which policy acts.

For a normalized expert action window $A^E$, we form
$A^\ell=\sqrt{\bar\alpha_\ell}A^E+\sqrt{1-\bar\alpha_\ell}\epsilon$,
with $\epsilon\sim\mathcal N(0,I)$, a uniformly sampled diffusion timestep $\ell$, and cumulative signal-retention coefficient $\bar\alpha_\ell$.
The student predicts clean actions through
\begin{equation}
 \mathcal L(\theta)=\mathbb E\!\left[
 \frac{\|M\odot(f_\theta(A^\ell,\ell,O,C)-A^E)\|_2^2}
 {\max(1,\sum_j M_j)}
 \right],
\label{eq:diffusion}
\end{equation}
where $f_\theta$ is the denoising network and $M$ masks unsupervised entries; $j$ ranges over all window timesteps and action coordinates.
The expectation covers training windows, diffusion timesteps, and noise.
% Both rollout branches share this objective.
\vspace{-3mm}
\paragraph{Relative robot conditioning.}
We encode reference motion relative to the torso to reduce
dependence on global placement while retaining orientation
feedback.
Let $(\bar{\mathbf{p}}_k,\bar{\mathbf{R}}_k)$ denote the
reference torso pose, and $s$ the conditioning window's
anchor timestep, with actual torso orientation $\mathbf{R}_s$.
We compute
\begin{equation}
\begin{aligned}
\Delta\mathbf{p}_k
&= \bar{\mathbf{R}}_s^\top
   (\bar{\mathbf{p}}_k-\bar{\mathbf{p}}_s),\\
\Delta\mathbf{R}_k
&= \mathbf{R}_s^\top\bar{\mathbf{R}}_k.
\end{aligned}
\label{eq:relative}
\end{equation}
Reference-relative displacement describes the intended
motion independently of its global placement, while
actual-relative orientation preserves feedback on rotational
tracking.
Together with reference joint positions and velocities,
these quantities guide the intended behavior, while
egocentric perception supports closed-loop interaction.
\paragraph{Multi-Process Distillation Acceleration.}
To accelerate visual rollout collection~\cite{zhuang2024parkourlearning,he2025viral}, we run parallel Isaac Sim workers that collect BC and DAgger rollouts with expert labels across reference groups.
A separate learner aggregates these data to update the shared student, whose checkpoint is synchronized across workers between collection rounds.
% \changliu{I did not find $\theta'$ in Fig 2.}

\subsection{Sim-to-Real Transfer}
\label{sec:sim2real}

After training and distillation in simulation, \projname{} is deployed in the real world without additional policy learning.
For video references, we estimate world-grounded SMPL~\cite{loper2015smpl} body motion with GVHMR~\cite{shen2024gvhmr}, and extract dynamic object motion with FoundationPose~\cite{wen2024foundationposeunified6dpose}. We retarget them with the same pipeline as MoCap data (Sec.~\ref{sec:experts}), 
so both sources share the same conditioning interface.
% Before execution, the reference is aligned to the robot's initial pose. 
% We use the G1 torso link as the student's base body,
% retaining its full roll, pitch, and yaw rather than a
% heading-only frame.
% At deployment in the real world, student inputs follow the same torso-frame
% convention as in simulation. We segment RGB images using Grounded-SAM~\cite{ren2024groundedsamassemblingopenworld}, prompted with the target object and robot body names.
% % , such as ``box, '' ``stairs, ''and ``robot bodies''.
% The masks select pixels from aligned depth images, which are back-projected into point clouds.
% During execution, the frozen student is queried with onboard proprioception, the latest segmented point clouds, and the current reference horizon.
Before execution, the reference is aligned to the robot's
initial pose.
We use \texttt{torso\_link} as the base body with full
roll, pitch, and yaw, and construct reference conditioning
as in Eq.~\eqref{eq:relative}, consistently with training.
We segment RGB images using
Grounded-SAM~2~\cite{ren2024groundedsamassemblingopenworld,ravi2024sam2}, which uses text grounding to prompt SAM~2 with the target object and robot body names.
The masks select pixels from aligned depth images, which
are back-projected into point clouds.
The frozen student combines onboard proprioception,
the latest segmented point clouds, and the reference
horizon to generate actions.

To facilitate transfer, we use task-specific reference and scene augmentation (ablated in Sec.~\ref{sec:abl_traj_aug}) and perturb visual observations to approximate real-sensor artifacts. Task-specific augmentation settings are detailed in the supplementary material.
% \changliu{it worth adding a figure to illustrate how the closed-loop tracking works. Also, we need to discuss why the robot needs the full reference trajectory, as there are many work that does real-time tracking, and it seems they are working well (but with human operator closing the loop). Here, we essentially want the robot to close the loop itself.}
% \changliu{define all these symbols}
% \jinxian{at the upper right of $\pi'$}

\section{Real-World Experiments}
\label{sec:realexp}

\paragraph{Experimental setup.}
Our experiments deploy a 29-DoF Unitree G1 humanoid equipped with boxing gloves. 
Perception is provided by an Intel RealSense D435i.
All policy inference runs on the robot's onboard NVIDIA Jetson Orin NX with 16 GB of memory.

We evaluate each task by having the robot follow the supplied reference trajectory. A trial succeeds if the robot completes the referenced motion without falling while maintaining the intended interaction.
\subsection{Results}
\label{sec:real_results}

\paragraph{MoCap data as reference trajectories.}
We use object-interaction references from OMOMO~\cite{li2023omomo},
climbing references from OmniRetarget~\cite{yang2025omniretarget},
and sitting references from AMASS~\cite{mahmood2019amass}.
The same frozen policy handles moving-object interactions,
such as box carrying (80\% success), as well as whole-body
support transitions in climbing (100\%) and sitting.
Figure~\ref{fig:real} illustrates these executions alongside
trash-can carrying, chair and suitcase pushing, and an
additional bucket-lifting demonstration. 

% Table~\ref{tab:real_success} summarizes hardware success counts
% using the same frozen student policy.

\begin{figure}[t]
\centering
\includegraphics[width=\linewidth]{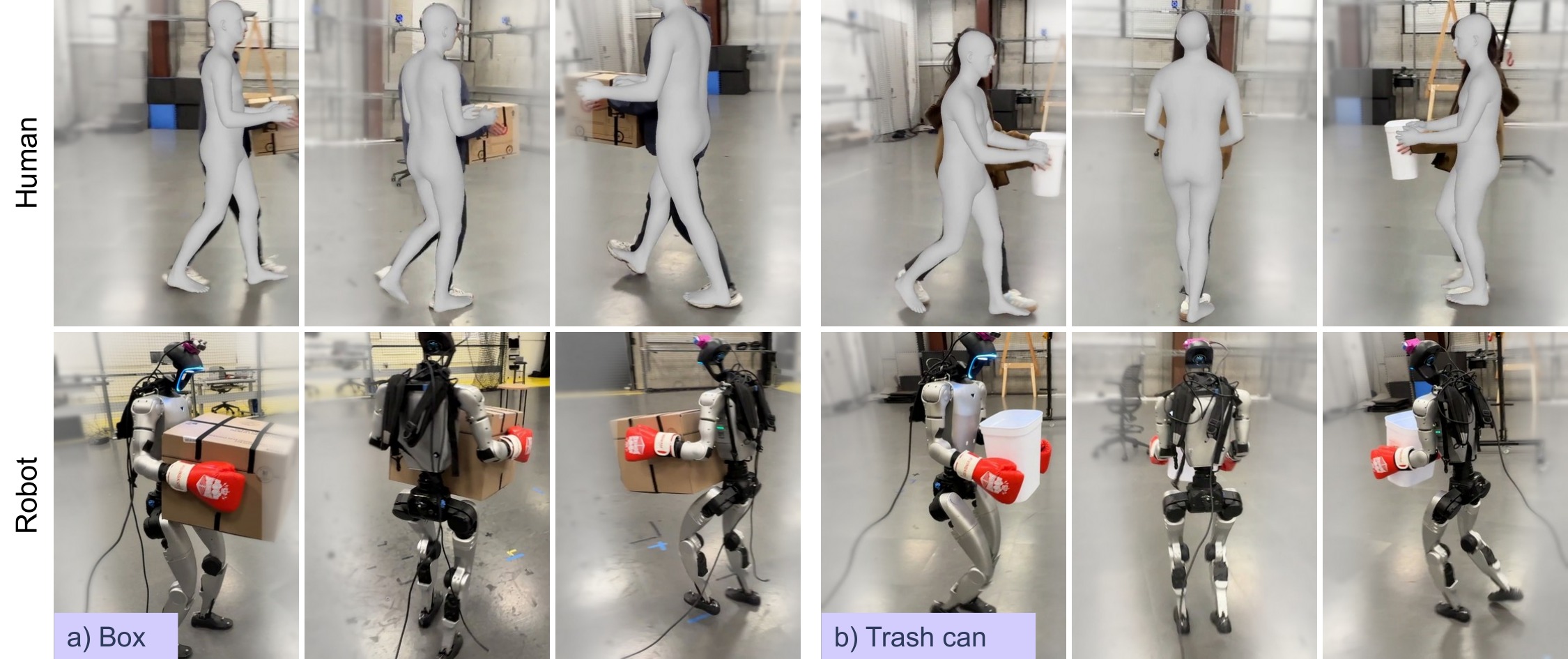}
% \vspace{-3mm}
\caption{\textbf{Following human-video references.} Top: human videos with the GVHMR-estimated human mesh overlaid. Bottom: \projname{} executing the video-derived references for (a) carrying a box and (b) carrying a trash can.}
\label{fig:real_transfer}
% \vspace{-3mm}
\end{figure}

% \subsection{Results}
% \label{sec:real_results}
% % The policy runs at 50 Hz using TensorRT FP16 with two DDIM denoising steps.
% % The inference latency is 13.6 ms.
% % \vspace{-2mm}
\paragraph{Video as reference trajectories.}
We record monocular RGB videos of human demonstrators in everyday scenes, and extract reference trajectories with GVHMR~\cite{shen2024gvhmr} as described in Sec.~\ref{sec:sim2real}.
We evaluate two tasks: \emph{Carry Box} and \emph{Carry Trash Can}, each requiring the robot to walk forward, turn around, and walk back while holding the object with both hands.
Figure~\ref{fig:real_transfer} shows the human videos and the corresponding robot executions: the robot reproduces the demonstrator's walking direction, turn, and two-handed object hold, using the same frozen student and conditioning interface as for MoCap references.

\subsection{Generalization}
\label{sec:real_generalization}

\begin{figure}[t]
\centering
\includegraphics[width=\linewidth]{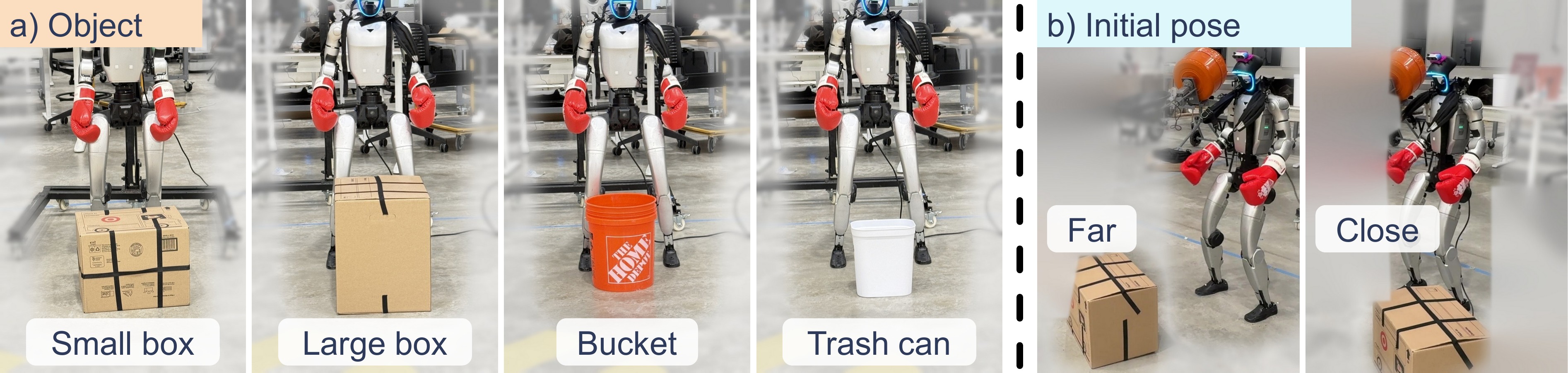}
% \vspace{-3mm}
\caption{\textbf{Generalization on hardware.} The same frozen policy handles (a) different objects and (b) different initial poses relative to the same box.}
\label{fig:real_generalization}
\vspace{-3mm}
\end{figure}

We test whether one frozen \projname{} policy handles scene changes that are not in the reference. 
Figure~\ref{fig:real_generalization}(a) varies the object
(two box sizes, a bucket, and a trash can), while
Figure~\ref{fig:real_generalization}(b) varies the robot's
initial position relative to the same box.
Each trial uses the same source robot-motion reference,
aligned to the robot's initial pose.
Object geometry is provided by onboard point clouds,
allowing the frozen policy to adapt to modest variations
within an observable and reachable range.
% Figure~\ref{fig:real_generalization}a varies the object: two box sizes, a bucket, and a trash can. Figure~\ref{fig:real_generalization}b varies where the robot starts relative to the same box. In each case the reference is aligned to the robot's initial pose, and the same policy completes the task. 
% \changliu{for different initial pose (far vs close), does the reference differ based on the object location or is it the same in the robot's frame?}

% \changliu{do the object and terrain need to remain the same as in the reference trajectory and in real?}
% \changliu{since this policy only uses the on-board sensors, I would suggest that you also get some footage of tasks executed in outdoor environments.}
% \changliu{and it would also be interesting to visualize numerically how the reference trajectory got mapped to the robot's own frame, how the policy tracking deviates from that reference, but finally be pulled back.}

\begin{table*}[t]
\centering
\caption{Simulation results across task families.}
\label{tab:main}
\label{tab:climbing_sitting}
\scriptsize
\setlength{\tabcolsep}{0.7pt}
\renewcommand{\arraystretch}{1.15}
\begin{tabularx}{\textwidth}{@{}l*{6}{>{\hsize=.70\hsize\linewidth=\hsize\centering\arraybackslash}X>{\hsize=1.08\hsize\linewidth=\hsize\centering\arraybackslash}X>{\hsize=1.22\hsize\linewidth=\hsize\centering\arraybackslash}X}@{}}
\toprule
 & \multicolumn{6}{c}{Object interaction} & \multicolumn{6}{c}{Climbing} & \multicolumn{6}{c}{Sitting}\\
\cmidrule(lr){2-7}\cmidrule(lr){8-13}\cmidrule(lr){14-19}
 & \multicolumn{3}{c}{Seen} & \multicolumn{3}{c}{Unseen} & \multicolumn{3}{c}{Seen} & \multicolumn{3}{c}{Unseen} & \multicolumn{3}{c}{Seen} & \multicolumn{3}{c}{Unseen}\\
\cmidrule(lr){2-4}\cmidrule(lr){5-7}\cmidrule(lr){8-10}\cmidrule(lr){11-13}\cmidrule(lr){14-16}\cmidrule(lr){17-19}
Method & {\fontsize{5.2}{6.2}\selectfont\mbox{Prog.\,$\uparrow$}} & {\fontsize{5.2}{6.2}\selectfont\mbox{$E_p$ (cm)\,$\downarrow$}} & {\fontsize{5.2}{6.2}\selectfont\mbox{$E_R$ (rad)\,$\downarrow$}} & {\fontsize{5.2}{6.2}\selectfont\mbox{Prog.\,$\uparrow$}} & {\fontsize{5.2}{6.2}\selectfont\mbox{$E_p$ (cm)\,$\downarrow$}} & {\fontsize{5.2}{6.2}\selectfont\mbox{$E_R$ (rad)\,$\downarrow$}} & {\fontsize{5.2}{6.2}\selectfont\mbox{Prog.\,$\uparrow$}} & {\fontsize{5.2}{6.2}\selectfont\mbox{$E_h$ (cm)\,$\downarrow$}} & {\fontsize{5.2}{6.2}\selectfont\mbox{$E_{\mathrm{tilt}}$ (rad)\,$\downarrow$}} & {\fontsize{5.2}{6.2}\selectfont\mbox{Prog.\,$\uparrow$}} & {\fontsize{5.2}{6.2}\selectfont\mbox{$E_h$ (cm)\,$\downarrow$}} & {\fontsize{5.2}{6.2}\selectfont\mbox{$E_{\mathrm{tilt}}$ (rad)\,$\downarrow$}} & {\fontsize{5.2}{6.2}\selectfont\mbox{Prog.\,$\uparrow$}} & {\fontsize{5.2}{6.2}\selectfont\mbox{$E_h$ (cm)\,$\downarrow$}} & {\fontsize{5.2}{6.2}\selectfont\mbox{$E_{\mathrm{tilt}}$ (rad)\,$\downarrow$}} & {\fontsize{5.2}{6.2}\selectfont\mbox{Prog.\,$\uparrow$}} & {\fontsize{5.2}{6.2}\selectfont\mbox{$E_h$ (cm)\,$\downarrow$}} & {\fontsize{5.2}{6.2}\selectfont\mbox{$E_{\mathrm{tilt}}$ (rad)\,$\downarrow$}}\\
\midrule
Retarget-only & 0.154 & \textbf{6.71} & \textbf{0.14} & 0.162 & \textbf{9.00} & \textbf{0.16} & 0.105 & 5.59 & 0.25 & 0.225 & 5.64 & 0.21 & 0.138 & 8.05 & 0.27 & 0.142 & 9.46 & 0.28 \\
OmniRetarget~\cite{yang2025omniretarget} & 0.436 & 12.20 & 0.27 & 0.269 & 15.20 & 0.32 & \cellcolor{progressbest}\textbf{0.542} & \cellcolor{progressbest}\textbf{3.17} & \cellcolor{progressbest}0.31 & 0.431 & 4.40 & 0.34 & 0.612 & \textbf{2.18} & 0.24 & 0.218 & 5.56 & 0.38 \\
SONIC~\cite{luo2025sonic} & 0.388 & 18.70 & 0.35 & 0.365 & 18.50 & 0.27 & 0.392 & 4.03 & \textbf{0.11} & 0.598 & \textbf{3.67} & \textbf{0.14} & \cellcolor{progressbest}\textbf{0.711} & \cellcolor{progressbest}3.83 & \cellcolor{progressbest}\textbf{0.16} & 0.527 & 5.80 & \textbf{0.17} \\
Shared offline BC & 0.443 & 11.72 & 0.29 & 0.259 & 13.55 & 0.30 & 0.411 & 3.81 & 0.25 & 0.541 & 4.49 & 0.41 & 0.287 & 4.12 & 0.25 & 0.311 & 3.57 & 0.25 \\
\projname{} & \cellcolor{progressbest}\textbf{0.849} & \cellcolor{progressbest}15.37 & \cellcolor{progressbest}0.26 & \cellcolor{progressbest}\textbf{0.828} & \cellcolor{progressbest}17.35 & \cellcolor{progressbest}0.36 & 0.472 & 4.74 & 0.25 & \cellcolor{progressbest}\textbf{0.631} & \cellcolor{progressbest}6.20 & \cellcolor{progressbest}0.35 & 0.567 & 3.37 & 0.23 & \cellcolor{progressbest}\textbf{0.557} & \cellcolor{progressbest}\textbf{3.29} & \cellcolor{progressbest}0.22 \\
\noalign{\vskip 2pt\hbox to \textwidth{\leaders\hbox{\rule{3pt}{0.4pt}\hskip 2pt}\hfill}\vskip 2pt}
Our RL experts & 0.872 & 10.60 & 0.20 & 0.848 & 11.80 & 0.18 & 0.559 & 2.83 & 0.20 & 0.570 & 2.67 & 0.28 & 0.844 & 4.00 & 0.22 & 0.716 & 4.54 & 0.22 \\
\bottomrule
\end{tabularx}
\end{table*}
\begin{figure*}[t]
\centering
\includegraphics[width=0.98\textwidth]{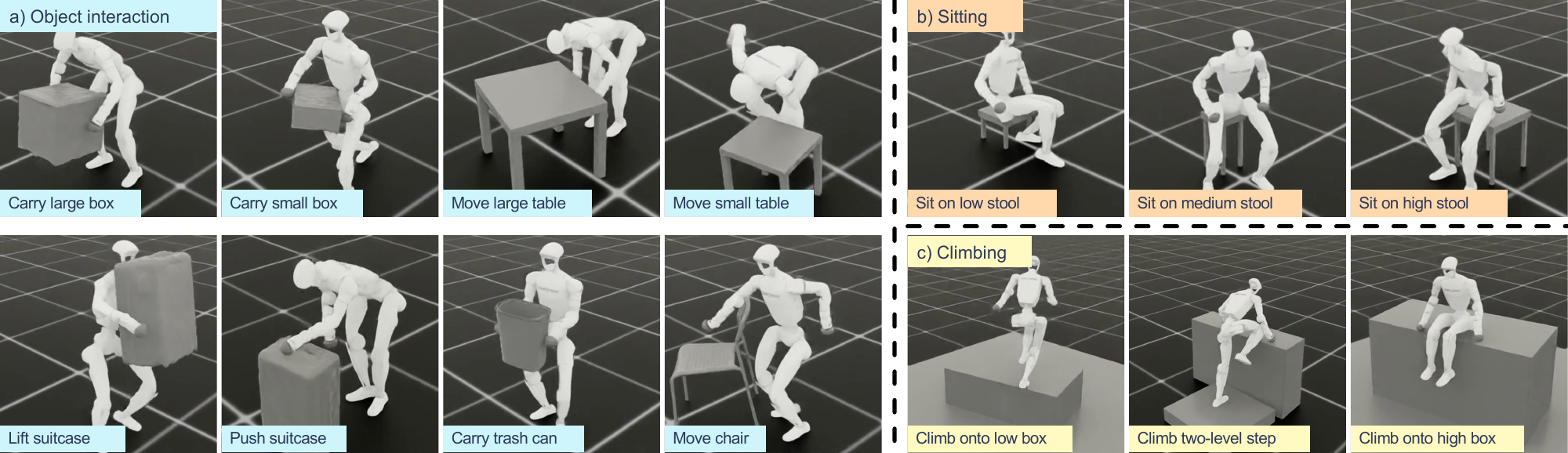}
\caption{\textbf{Simulation task suite.} Representative Isaac Sim environments used for training and evaluation, executed by \projname{} on a G1. (a) Object interaction with objects of different shapes and sizes (boxes, tables, suitcases, a trash can, and a chair); (b) sitting on low, medium, and high stools; (c) climbing onto a low box, a two-level step, and a high box.}
\label{fig:sim_tasks}
\end{figure*}
\section{Simulation Experiments}
\label{sec:experiment}
\label{sec:simexp}

\paragraph{Experimental setup.}
We evaluate the G1 policy in Isaac Sim on representative retargeted trajectories spanning object interaction, sitting, and climbing (Fig.~\ref{fig:sim_tasks}). The suite covers carrying, lifting, pushing, and moving objects of varied geometry, sitting on stools of different heights, and climbing onto steps and boxes with different heights and footprints. 
The evaluation set includes 40 seen and 30 held-out OMOMO reference trajectories~\cite{li2023omomo} for object interaction, 10 seen and 7 held-out climbing trajectories from OmniRetarget~\cite{yang2025omniretarget}, and 7 seen and 4 held-out AMASS-based sitting trajectories~\cite{mahmood2019amass}. 
% \changliu{what does the last sentence mean?}
% Episodes start at the beginning of each reference.
% \changliu{need to discuss hardware usage, training time, inference time, etc.}
Training all privileged RL experts requires 64 GPU-hours on NVIDIA RTX 3090 GPUs. Visual distillation runs for 24 hours using one NVIDIA RTX A4000 and one RTX A6000, totaling 48 GPU-hours. 

\subsection{Baselines and Results}
\label{sec:sim_baselines}

\paragraph{Baselines.}
We compare against the following baselines:
\textit{Retarget-only} directly uses the retargeted motion as joint-position targets without learned dynamics compensation.
\textit{OmniRetarget}~\cite{yang2025omniretarget} uses a tracking-based RL method that we train following the original method on the same training references used by our approach.
\textit{SONIC}~\cite{luo2025sonic} is a pretrained humanoid motion-tracking foundation model without object observations.
We evaluate its original-release Default G1 policy with frozen encoder and decoder, conditioned on the same robot-motion references.
\textit{Shared offline BC} uses the same student architecture and observation interface as \projname{}, but is trained on a fixed dataset of expert rollouts collected once, without subsequent online collection.
Finally, \textit{our RL experts} (Sec.~\ref{sec:experts}) provide a reference for performance with privileged state observations.
\vspace{-3mm}

% % \changliu{can the execution be faster than the reference, i.e., completing a 30s task in 15s?}

\paragraph{Metrics.}
We evaluate performance using:
1) Progress, quantified by the fraction of the reference
trajectory completed before failure;
2) object translation error $E_p$, quantified by the mean
Euclidean distance between actual and reference object positions;
3) object rotation error $E_R$, quantified by the mean angular
difference between actual and reference object orientations;
4) root height tracking error $E_h$, quantified by the mean
absolute deviation from the reference root height;
and 5) root tilt tracking error $E_{\mathrm{tilt}}$, quantified
by the mean angle between actual and reference root up axes,
excluding pure heading differences.
Progress measures overall trajectory completion across all tasks.
For moving-object interaction, $E_p$ and $E_R$ assess
object-motion fidelity. For climbing and sitting, $E_h$ and
$E_{\mathrm{tilt}}$ assess postural tracking relevant to balance.
% Higher Progress and lower errors are better.

Failure is detected when any of the following conditions
holds for three consecutive steps: object height error
exceeds $25$\,cm, object rotation error exceeds $\pi/2$\,rad,
root height error exceeds $30$\,cm, or the norm of the
difference between actual and reference projected gravity
exceeds $0.8$.
Object-related criteria apply only to moving-object interactions.
Progress is computed up to the first detected failure,
or equals one if the full reference is completed without failure.
Tracking errors are averaged over frames before failure
within each episode and then across evaluation episodes.
\begin{figure}[t]
\centering
\includegraphics[width=0.96\linewidth]{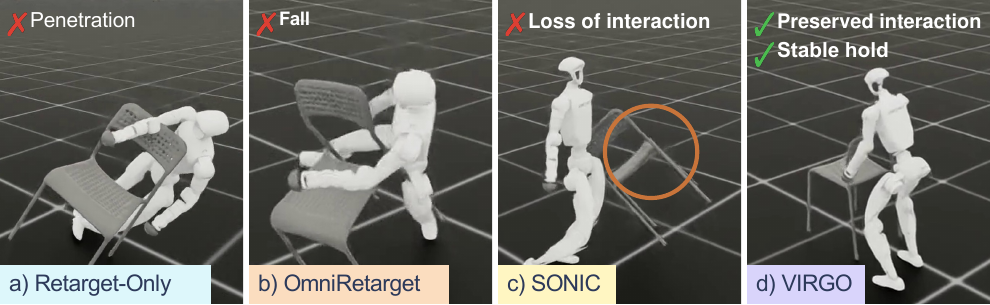}
\caption{\textbf{Comparison on an unseen lift-chair reference.} Retarget-only penetrates, OmniRetarget falls, SONIC loses the interaction, while \projname{} maintains a stable hold.}
\label{fig:sim_compare}
\vspace{-3mm}
\end{figure}

\vspace{-3mm}\paragraph{Results.}
Table~\ref{tab:main} summarizes the simulation results.
% \textit{Retarget-only} has low Progress across all three task families, and the qualitative example in Fig.~\ref{fig:sim_compare} shows object penetration. \textit{OmniRetarget}'s Progress decreases from seen to held-out references in each family, most sharply for sitting, but this comparison alone does not isolate source-motion generalization. \textit{SONIC} is competitive on sitting but trails \projname{} in moving-object interaction and climbing Progress. \textit{Shared offline BC} also trails \projname{} in object-interaction Progress, consistent with the value of student-state coverage during distillation.
% \projname{} approaches our privileged RL experts in object-interaction Progress on both splits and exceeds the evaluated deployable baselines on that metric. 
% Its object-pose errors remain higher than those of several
% baselines.
\textit{Retarget-only} has low Progress due to physically
implausible execution, while \textit{OmniRetarget} tracks
seen references well but generalizes less effectively to
held-out motions. \textit{SONIC} is competitive on sitting, whereas
\textit{Shared offline BC} lags behind \projname{} in
object-interaction Progress, consistent with its limited
student-state coverage. Although \projname{} trails \textit{SONIC} on seen sitting,
it exceeds \textit{SONIC} on held-out sitting and both
climbing splits. Its higher object-pose errors should be interpreted alongside
Progress, since errors exclude frames after failure.

Figure~\ref{fig:sim_compare} illustrates its ability to maintain
the intended interaction despite baseline failures.
Overall, our policy achieves the highest held-out Progress
across all three task families among the evaluated deployable
methods, with particularly substantial gains in object interaction.
\subsection{Trajectory Randomization and Augmentation}
\label{sec:abl_traj_aug}

We train two large-box RL experts with the same data and training budget, with and without trajectory randomization, and evaluate on the same batch of randomized trajectories.
As shown in Table~\ref{tab:traj_rand}, trajectory randomization substantially improves both Progress and success rate, with lower tracking errors. The expert stays reliable when the robot and object deviate from the nominal reference, which is where the student needs its labels during distillation.
\begin{table}[t]
\centering
\caption{Trajectory randomization for RL experts.}
\label{tab:traj_rand}
\vspace{-2mm}
\scriptsize
\setlength{\tabcolsep}{4pt}
\renewcommand{\arraystretch}{1.15}
\begin{tabular}{lcccc}
\toprule
Expert training & {\fontsize{6}{7}\selectfont Prog.\,$\uparrow$} & {\fontsize{6}{7}\selectfont Succ. (\%)\,$\uparrow$} & {\fontsize{6}{7}\selectfont $E_p$ (cm)\,$\downarrow$} & {\fontsize{6}{7}\selectfont $E_{\mathrm{tilt}}$ (rad)\,$\downarrow$} \\
\midrule
With traj. randomization & \textbf{0.849} & \textbf{77.4\%} & \textbf{9.9} & \textbf{0.153} \\
Without traj. randomization & 0.521 & 34.4\% & 10.5 & 0.207 \\
\bottomrule
\end{tabular}
\vspace{-2mm}
\end{table}

\begin{figure}[t]
    \centering
    \includegraphics[width=0.98\linewidth]{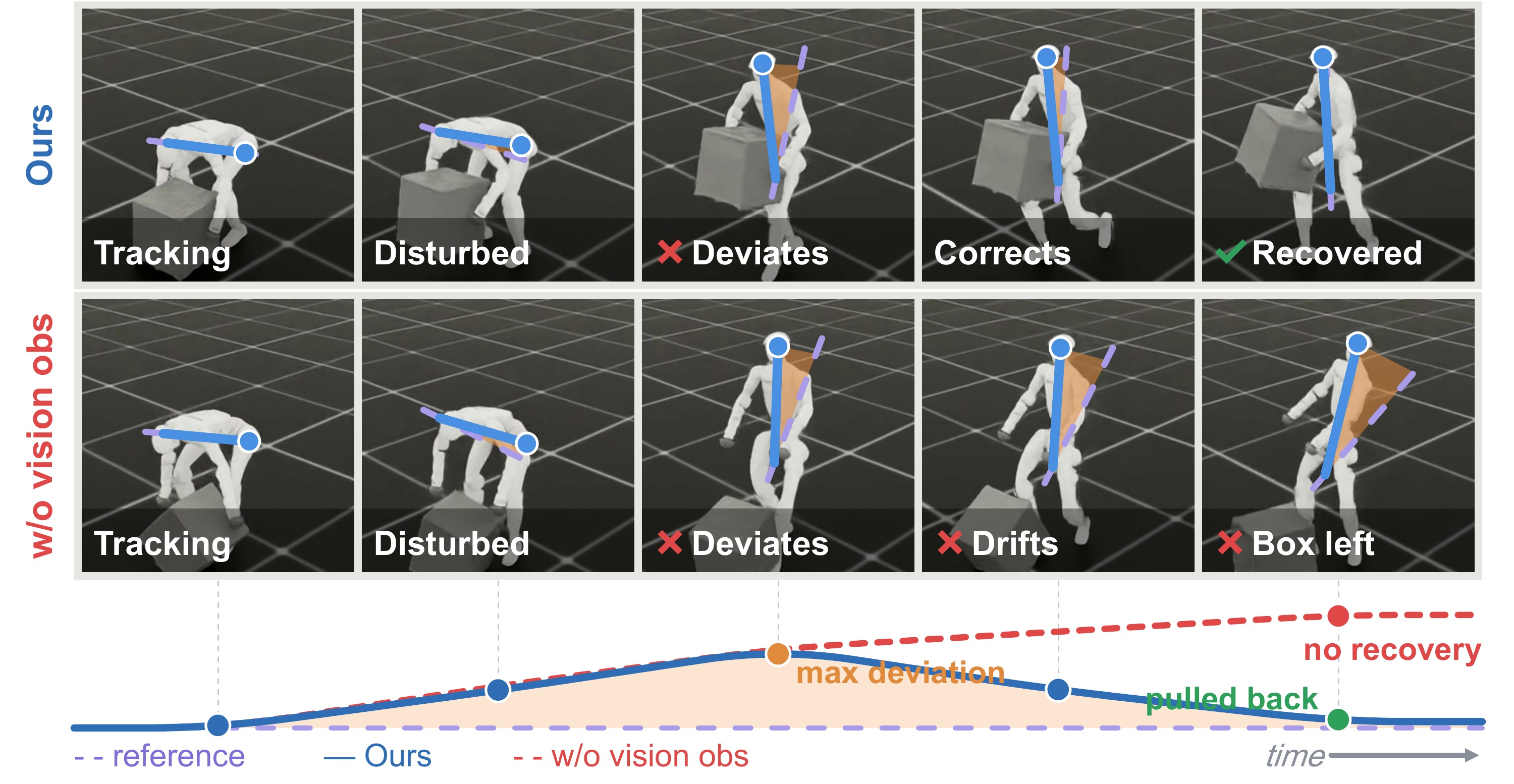}
    \caption{\textbf{Recovery comparison.}
After a disturbance, visual feedback enables recovery toward the dashed reference, and the shaded areas indicate deviations in the schematic curves.}
    \label{fig:recovery}
\end{figure}
\subsection{Visual Observations}

Visual observations provide feedback on robot--object geometry,
enabling closed-loop recovery when execution drifts from the
reference.
Figure~\ref{fig:recovery} compares \projname{} with its ablated variant without visual
observations.
With vision, the policy corrects the drift and resumes the
intended interaction; without vision, the robot continues
tracking the reference body motion while leaving the fallen
box behind.

\subsection{DAgger versus BC Visual Distillation}
Following VIRAL~\cite{he2025viral}, we study the balance between expert supervision and student-state coverage by varying the fraction of BC (expert-only) environments, $\rho\in\{0,0.3,0.5,0.7,1\}$. For $0<\rho<1$, DAgger environments follow the expert-mixing schedule in Sec.~\ref{sec:distill}. At $\rho=0$, we additionally set $\beta_e=0$ to obtain pure student-rollout DAgger; this endpoint changes both the BC allocation and expert intervention. The $\rho=1$ endpoint gives pure BC. Unlike the shared offline BC baseline, the $\rho=1$ setting in our mixture ablation refreshes expert-only rollouts at every collection round; the two therefore differ in their training-data collection protocol. 

\begin{figure}[t]
\centering
\includegraphics[width=0.98\linewidth]{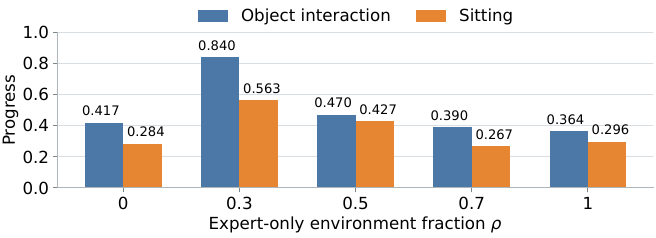}
\vspace{-2mm}
\caption{\textbf{DAgger\&BC mixture.} Progress vs. expert-only fraction $\rho$, from pure DAgger ($\rho=0$) to pure BC ($\rho=1$). The mixture peaks at $\rho=0.3$.}
\label{fig:bc_dagger_ablation}
\vspace{-3mm}
\end{figure}

Figure~\ref{fig:bc_dagger_ablation} reports closed-loop Progress on object interaction and sitting. The mixture peaks at $\rho=0.3$ for both tasks and clearly outperforms both pure DAgger and pure BC. 
% Training on student-visited states also teaches recovery: when the student drifts from the reference, it is pulled back rather than diverging (Fig.~\ref{fig:recovery}).

\begin{figure}[t]
\centering
\includegraphics[width=0.98\linewidth]{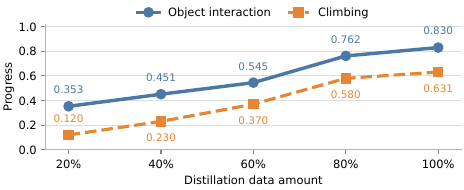}
\caption{\textbf{Distillation data scaling.} Progress on object interaction and climbing versus the fraction of distillation data used.}
\label{fig:ablation_curves}
\end{figure}

\subsection{Distillation Data Scaling}
We study how the amount of expert-labeled distillation data affects the shared policy, using 20\%, 40\%, 60\%, 80\%, and 100\% of the full training set. 
The expert bank, student architecture, and evaluation trajectories remain fixed across budgets. Figure~\ref{fig:ablation_curves} reports Progress separately for object interaction and climbing, measuring the benefit of additional supervision under the same evaluation protocol.
% \changliu{bold the best performing entries}
% \changliu{why unseen scenarios almost have similar performance as seen scenarios in object interaction, and slightly better in climbing?}

\section{Conclusion}
% We present \projname{}, which learns a single reference-conditioned
% visuomotor policy for diverse whole-body interactions from human
% demonstrations.
% Our expert-to-generalist framework consolidates motion-group RL
% experts into a shared policy, combining reference guidance with
% onboard visual feedback for autonomous closed-loop execution.
% The same frozen policy transfers zero-shot to a physical humanoid,
% providing a practical route from human demonstrations to diverse
% robot interactions without teleoperated robot training data or
% task-specific fine-tuning.
In this work, we addressed the challenge of learning autonomous visual
closed-loop control across diverse humanoid interactions
without costly teleoperated robot demonstrations.
We introduce \projname{}, an expert-to-generalist framework
that distills motion-group RL experts trained on human
demonstrations into one reference-conditioned visuomotor policy.
The student achieves a mean trajectory completion ratio of
0.828 on held-out object-interaction references, outperforming
the evaluated deployable baselines.
The same frozen policy transfers zero-shot to a physical
humanoid, executing unseen MoCap and video-derived references
across object interactions and whole-body support transitions.
% \changliu{if time allows, we should also try Supermimic for the expert.}
\vspace{-2mm}
\paragraph{Limitations and future work.}
Although \projname{} uses visual feedback for closed-loop execution, it follows a supplied motion reference and may not always recover when substantial deviations require a different interaction.
% Its generalization is constrained by the coverage of training demonstrations.
Future work will expand demonstration diversity and improve reference adaptation for more varied whole-body interactions.
% The iterative denoising of the diffusion policy introduces inference latency, and segmenting depth-derived point clouds adds cumbersome preprocessing for real-world deployment.
% Future work will explore lightweight policy architectures, such as compact MLP-based students, and more direct RGB or RGB-D observations with learned visual encoders.
% \changliu{need to report latency in the result section.}

% Required by CVPR 2027; excluded from the main-text page limit.
\section*{LLM Usage Statement}
AI tools, including OpenAI Codex, were used to assist with manuscript review, grammar and clarity checks, figure adjustment, and LaTeX formatting corrections. 
The authors reviewed and verified the final text, citations, figures, and scientific claims, and remain responsible for the content of this paper.

{\small\bibliographystyle{ieeenat_fullname}\bibliography{main}}
\clearpage
\setcounter{page}{1}
\setcounter{section}{0}
\setcounter{figure}{0}
\setcounter{table}{0}
\setcounter{equation}{0}
\renewcommand{\thesection}{\Alph{section}}
\renewcommand{\thefigure}{S\arabic{figure}}
\renewcommand{\thetable}{S\arabic{table}}
\renewcommand{\theequation}{S\arabic{equation}}
\maketitlesupplementary
\raggedbottom
\renewcommand{\dbltopfraction}{0.95}
\renewcommand{\textfraction}{0.05}
\setcounter{dbltopnumber}{3}

\section{RL Training Details}
\label{sec:rl}\label{sec:supp_training}
\subsection{Simulation and Control}
We train G1 tracking experts with PPO~\cite{schulman2017ppo} in Isaac Sim. Each action specifies a joint target $q_j^{\mathrm{des}}=q_j^0+s_ja_j$, where $q^0$ is the default posture and $s_j=0.25\tau_j^{\max}/K_{p,j}$. PD control applies $\tau_j=K_{p,j}(q_j^{\mathrm{des}}-q_j)-K_{d,j}\dot q_j$, clipped to the torque limit $\tau_j^{\max}$.

\par\smallskip
\noindent\begin{minipage}{\linewidth}
\captionsetup{hypcap=false}
\centering\small
\captionof{table}{Simulation and PPO parameters. O/C/S denote object interaction, climbing, and sitting.}
\label{tab:supp_rl_config}
\setlength{\tabcolsep}{3pt}
\begin{tabularx}{\linewidth}{@{}Xl@{}}
\toprule
Parameter & Value \\
\midrule
Physics / control rate & 200 / 50\,Hz \\
Environments (O / C,S) & 4096 / 1024 \\
Environment spacing (O / C,S) & 2.5 / 5\,m \\
PhysX position / velocity iterations & 8 / 4 \\
Substeps / episode limit & 1 / 10\,s \\
Actor / critic hidden widths & $(512,256,128)$ \\
Activation / observation normalization & ELU / running statistics \\
Optimizer / initial learning rate & AdamW / $10^{-3}$ \\
Weight decay / target KL & 0 / 0.01 (adaptive LR) \\
Rollout steps / minibatches / PPO epochs & 24 / 4 / 5 \\
PPO clipping / discount / GAE & 0.2 / 0.99 / 0.95 \\
Entropy / value-loss coefficient & 0.005 / 1 \\
Gradient norm limit / initial action std. & 1 / 1 \\
Training iterations (O / C / S) & 8--28k / 4--12k / 4.5--7.5k \\
Seed (O / C / S) & 42 / 1 / 42 \\
Total expert-training cost & 64 RTX 3090 GPU-hours \\
\bottomrule
\end{tabularx}
\end{minipage}\par\smallskip

\par\smallskip
\noindent\begin{minipage}{\linewidth}
\captionsetup{hypcap=false}
\centering\small
\captionof{table}{PD gains and torque limits (Nm); shared by left/right joints.}
\label{tab:supp_pd}
\setlength{\tabcolsep}{3pt}
\begin{tabularx}{\linewidth}{@{}Xrrr@{}}
\toprule
Joint group & $K_p$ & $K_d$ & $\tau^{\max}$ \\
\midrule
Hip pitch/yaw; waist yaw & 40.179 & 2.558 & 88 \\
Hip roll; knee & 99.098 & 6.309 & 139 \\
Ankle; waist pitch/roll & 28.501 & 1.814 & 50 \\
Shoulder; elbow; wrist roll & 14.251 & 0.907 & 25 \\
Wrist pitch/yaw & 16.778 & 1.068 & 5 \\
\bottomrule
\end{tabularx}
\end{minipage}\par\smallskip
\subsection{Observations}
The actor uses the terms in Table~\ref{tab:supp_observation}; the critic adds privileged states. Actor noise is uniform in $[-\eta,\eta]$, and critic noise is disabled. Both use one-step observations clipped to $[-100,100]$. The $B=14$ tracked bodies are the pelvis, torso, and bilateral hip-roll, knee, ankle-roll, shoulder-roll, elbow, and wrist-yaw links.

\par\smallskip
\noindent\begin{minipage}{\linewidth}
\captionsetup{hypcap=false}
\centering\small
\captionof{table}{Expert inputs. Orientations use a 6D representation.}
\label{tab:supp_observation}
\setlength{\tabcolsep}{3pt}
\begin{tabularx}{\linewidth}{@{}Xrr@{}}
\toprule
Actor and critic & Dim. & $\eta$ \\
\midrule
Reference joint positions / velocities & 58 & 0 \\
Reference torso orientation in robot torso frame & 6 & 0.05 \\
Base angular velocity & 3 & 0.2 \\
Joint positions relative to $q^0$ / velocities & 58 & 0.01 / 0.5 \\
Previous action & 29 & 0 \\
\midrule
\multicolumn{3}{@{}l}{Additional critic inputs (no noise)} \\
Reference torso position in robot torso frame & 3 & \\
Body positions / orientations in torso frame & 126 & \\
Base linear velocity & 3 & \\
Object pose / velocity features (O only) & 12 & \\
\midrule
Actor / critic total (O; C,S) & \multicolumn{2}{r@{}}{154 / (298; 286)} \\
\bottomrule
\end{tabularx}
\end{minipage}\par\smallskip
\subsection{Reward Terms}
The reward combines body tracking, moving-object tracking, and desired contacts with control regularization:
\begin{equation}
r_t=\Delta t\bigl(r_t^{\mathrm{body}}+m_o r_t^{\mathrm{obj}}
+r_t^{\mathrm{contact}}-r_t^{\mathrm{reg}}\bigr),
\label{eq:supp_reward}
\end{equation}
where $\Delta t=0.02$\,s and $m_o$ is one for moving-object interaction and zero otherwise. Bars denote reference quantities. We use $\phi_\sigma(e)=\exp(-e/\sigma^2)$ for squared deviations $e$, and $d_R(R,\bar R)$ for the relative rotation angle. The weights and scales are listed in Table~\ref{tab:supp_rewards}.

\paragraph{Body tracking.}
Let $r$ denote the torso and $b$ index the $B=14$ tracked bodies. The six tracking terms are
\begin{equation}
\begin{aligned}
r_t^{\mathrm{body}}={}&w_p\phi_{\sigma_p}(\|p_r-\bar p_r\|^2)\\
&+w_R\phi_{\sigma_R}(d_R(R_r,\bar R_r)^2)\\
&+w_{bp}\phi_{\sigma_{bp}}\!\left(\frac1B\sum_b\|p_b-\tilde p_b\|^2\right)\\
&+w_{bR}\phi_{\sigma_{bR}}\!\left(\frac1B\sum_b d_R(R_b,\tilde R_b)^2\right)\\
&+w_v\phi_{\sigma_v}\!\left(\frac1B\sum_b\|v_b-\bar v_b\|^2\right)\\
&+w_\omega\phi_{\sigma_\omega}\!\left(\frac1B\sum_b\|\omega_b-\bar\omega_b\|^2\right).
\end{aligned}
\label{eq:supp_body_reward}
\end{equation}
Here $p,R,v,\omega$ denote position, orientation, and world-frame linear and angular velocity. The body targets $(\tilde p_b,\tilde R_b)$ use the randomized reference when augmentation is enabled. Otherwise, they are aligned to the robot's planar position and yaw while retaining reference height.

\paragraph{Object and contact tracking.}
For an interacting moving object $o$, we reward its pose tracking:
\begin{equation}
\begin{aligned}
r_t^{\mathrm{obj}}={}&w_{op}\phi_{\sigma_{op}}(\|p_o-\bar p_o\|^2)\\
&+w_{oR}\phi_{\sigma_{oR}}(d_R(R_o,\bar R_o)^2).
\end{aligned}
\label{eq:supp_object_reward}
\end{equation}
The contact reward counts desired contacts realized in simulation:
\begin{equation}
r_t^{\mathrm{contact}}=w_c\sum_{b,k}\bar c_{bk}\,
\mathbb{1}[F_{bk}>1\,\mathrm{N}],
\label{eq:supp_contact_reward}
\end{equation}
where $\bar c_{bk}$ is the reference contact label between body part $b$ and scene target $k$, and $F_{bk}$ is their measured force magnitude. Only desired-positive contacts contribute; no separate contact-mismatch penalty is applied.

\paragraph{Regularization.}
We penalize action variation, soft joint-limit violation, and undesired contacts:
\begin{equation}
\begin{aligned}
r_t^{\mathrm{reg}}={}&w_a\|a_t-a_{t-1}\|^2\\
&+w_q\sum_j\bigl([q_j^- -q_j]_+ +[q_j-q_j^+]_+\bigr)\\
&+w_u\sum_{b\in\mathcal U}\mathbb{1}\!\left[
\max_{\tau\in\mathcal H}\|F_{b,\tau}\|>1\,\mathrm{N}\right].
\end{aligned}
\label{eq:supp_regularization}
\end{equation}
The soft bounds $q_j^-,q_j^+$ span 90\% of each physical joint range about its midpoint, and $[x]_+=\max(x,0)$. The undesired-contact set $\mathcal U$ excludes the feet, wrist-yaw links, and ankle-roll links; $\mathcal H$ is the force-history window. We set $w_u=0$ for sitting.

\par\medskip
\noindent\begin{minipage}{\linewidth}
\captionsetup{hypcap=false}
\centering\small
\captionof{table}{Reward weights and tracking scales. Regularization weights are positive here and subtracted in Eq.~\eqref{eq:supp_reward}.}
\label{tab:supp_rewards}
\begin{tabularx}{\linewidth}{@{}Xrr@{}}
\toprule
Term & Weight & Scale $\sigma$ \\
\midrule
Torso position / orientation & 0.5 / 0.5 & 0.3 / 0.4 \\
Body position / orientation & 1 / 1 & 0.3 / 0.4 \\
Linear / angular velocity & 1 / 1 & 1 / 3.14 \\
Object position / orientation & 1 / 1 & 0.3 / 0.4 \\
Contact $w_c$ & 0.1 & --- \\
Action variation $w_a$ & 0.1 & --- \\
Joint limits $w_q$ & 10 & --- \\
Undesired contact $w_u$ (O,C / S) & 0.1 / 0 & --- \\
\bottomrule
\end{tabularx}
\end{minipage}\par\medskip

\subsection{Initialization and Termination}
Object/climbing experts use adaptive reference-phase initialization; sitting starts at frame zero. Table~\ref{tab:supp_termination} gives training failure thresholds. Unit gravity vectors $g,\bar g$ are expressed in the actual/reference torso frames. Sitting checks torso-relative offsets of the pelvis, knees, elbows, wrists, and ankles. Failures require three consecutive steps; non-finite errors terminate immediately.

\par\smallskip
\noindent\begin{minipage}{\linewidth}
\captionsetup{hypcap=false}
\centering\small
\captionof{table}{Training termination thresholds; distinct from evaluation metrics.}
\label{tab:supp_termination}
\setlength{\tabcolsep}{3pt}
\begin{tabularx}{\linewidth}{@{}Xrl@{}}
\toprule
Failure quantity & Threshold & Task \\
\midrule
Torso height / wrist--ankle target height error & 0.50 / 0.25\,m & O,C \\
Gravity projection error $|g_z-\bar g_z|$ & 0.80 & O,C \\
Object translation / rotation error & 0.25\,m / 0.80\,rad & O \\
Torso height / planar-position error & 0.25 / 0.50\,m & S \\
Gravity-vector error $\|g-\bar g\|$ & 0.80 & S \\
Maximum torso-relative body-offset error & 0.30\,m & S \\
\bottomrule
\end{tabularx}
\end{minipage}\par\smallskip
\section{Distillation Details}
\label{sec:supp_distillation}
\paragraph{Architecture and optimization.}
The student combines PointNet++~\cite{qi2017pointnetpp}, separate proprioception/reference MLPs, and a conditional temporal diffusion U-Net~\cite{chi2023diffusion,ze2024dp3}. Table~\ref{tab:supp_distill_config} summarizes the parameters.

\par\smallskip
\noindent\begin{minipage}{\linewidth}
\captionsetup{hypcap=false}
\centering\small
\captionof{table}{Student parameters.}
\label{tab:supp_distill_config}
\setlength{\tabcolsep}{3pt}
\begin{tabularx}{\linewidth}{@{}Xl@{}}
\toprule
Parameter & Value \\
\midrule
Point input (object tasks) & 2048 XYZ points \\
Proprioception / reference dimension & 93 / 67 \\
Proprioception / reference MLP widths & $(64,64)$ / same \\
Observation history & 2 \\
Action horizon / executed steps & 8 / 1 \\
Sparse reference offsets & 10, 20 steps \\
U-Net channels / kernel / norm groups & $(256,512,1024)$ / 5 / 8 \\
Diffusion embedding / conditioning & 128 / FiLM \\
Training diffusion steps / schedule & 100 / squared-cosine \\
Prediction target & Clean action \\
AdamW learning rate / weight decay & $8\times10^{-5}$ / $10^{-6}$ \\
AdamW betas / epsilon & $(0.95,0.999)$ / $10^{-8}$ \\
Batch size / seed & 32 / 42 \\
Simulation / hardware denoising steps & 10 / 2 \\
\bottomrule
\end{tabularx}
\end{minipage}\par\smallskip
\paragraph{Collection schedule.}
Distillation runs for 250 rounds with four collectors, 40 environments per active group, and a target of 1000 collection steps per group per round. Each round trains once on fresh expert-labeled rollouts without replacement; the selected model is from round 200. For zero-indexed round $e$, the BC fraction $\rho_e$ and teacher-mix expert probability $\beta_e$ are
\begin{equation}
(\rho_e,\beta_e)=
\begin{cases}
(0.3,\max(0.3,0.98^e)),&e<150,\\
(0,0.7\cdot0.98^e),&e\geq150.
\end{cases}
\label{eq:supp_mix_schedule}
\end{equation}
Conditions use source robot-motion references, without hindsight relabeling.

\paragraph{Augmentation.}
Trajectory yaw/translation perturbations are uniform within Table~\ref{tab:supp_randomization}'s bounds. Inverse kinematics updates the joint targets after geometric perturbations. Sitting collection keeps each fitted motion--stool pair fixed, with stool scaling, trajectory randomization, and initial-pose noise disabled, and resets to frame zero.

\par\smallskip
\noindent\begin{minipage}{\linewidth}
\captionsetup{hypcap=false}
\centering\small
\captionof{table}{Randomization. O/C: object/climbing RL; D: object distillation. Translation bounds/std. are in meters; yaw is in radians.}
\label{tab:supp_randomization}
\setlength{\tabcolsep}{3pt}
\begin{tabularx}{\linewidth}{@{}>{\raggedright\arraybackslash}p{0.27\linewidth}Xl@{}}
\toprule
Component & Range / setting & Stage \\
\midrule
Global transform & yaw $\pm0.05$; xyz $\pm(0.02,0.02,0)$ & O,C,D \\
Interaction transform & yaw $\pm0.1$; xyz $\pm(0.04,0.04,0)$ & O,C,D \\
Root / ankle transform & yaw $\pm0.015$; xyz $\pm(0.005,0.005,0)$ & O,C,D \\
Original fraction & 0.05 & O,C,D \\
Object scale / mass / friction & $[0.9,1.1]$ / $[0.9,1.1]$ / $[0.8,1.2]$ & O \\
Terrain scale & $[0.95,1.05]$, bottom-anchored & C \\
PC augmentation & prob. 0.5; jitter/translation std. 0.003/0.005 & D \\
PC scale / dropout & $[0.98,1.02]$ / 0.05 & D \\
Camera extrinsics; state/condition noise & Disabled & D \\
\bottomrule
\end{tabularx}
\end{minipage}\par\smallskip
\section{Motion Retargeting}
\label{sec:supp_retarget}
We adapt the interaction-mesh pipeline of OmniRetarget~\cite{yang2025omniretarget} to the G1. Human keypoints and sampled scene points form a tetrahedral mesh whose Laplacian coordinates encode local body--scene relationships. At each frame, sequential constrained optimization matches these coordinates while respecting joint limits, scene non-penetration, and stance-foot constraints. The previous frame initializes the solution, and temporal regularization reduces abrupt motion changes. We convert the result to a common 50\,Hz reference containing joint states, body poses and velocities, scene assets, and contact annotations.

\paragraph{Object interaction.}
For OMOMO~\cite{li2023omomo}, we retarget the human motion together with the moving-object mesh and pose trajectory. Human keypoints and object surface samples are expressed in the object's frame when constructing the interaction mesh. The robot configuration adapts to the prescribed object motion, preserving the task-relevant spatial relationships across embodiments. The exported reference includes both robot and object trajectories (Figure~\ref{fig:supp_retarget}).

\par\smallskip
\noindent\begin{minipage}{\linewidth}
\captionsetup{hypcap=false}
\centering
\includegraphics[height=2.9cm,trim=735bp 600bp 395bp 340bp,clip]{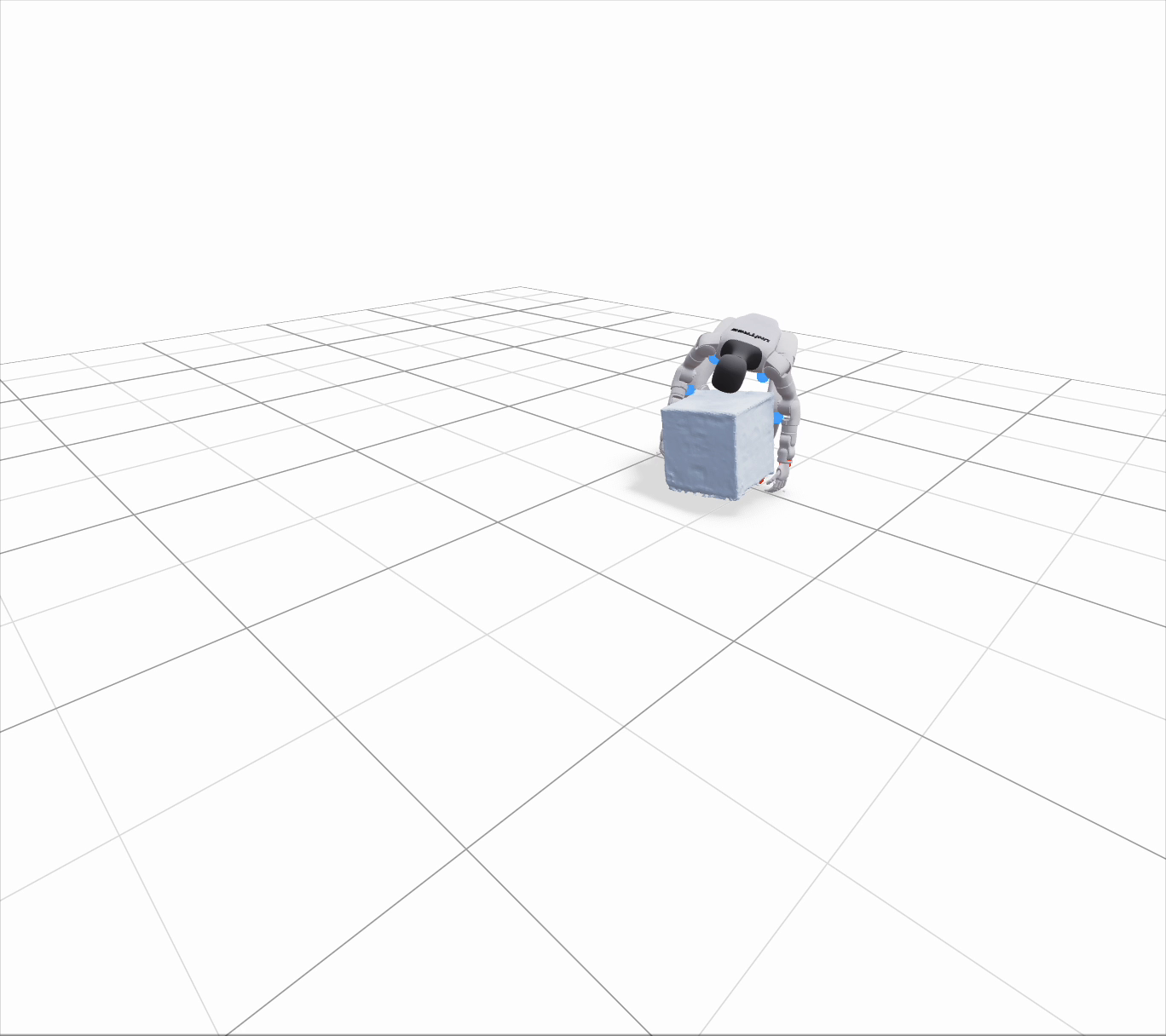}\hspace{5mm}
\includegraphics[height=2.9cm,trim=585bp 620bp 535bp 150bp,clip]{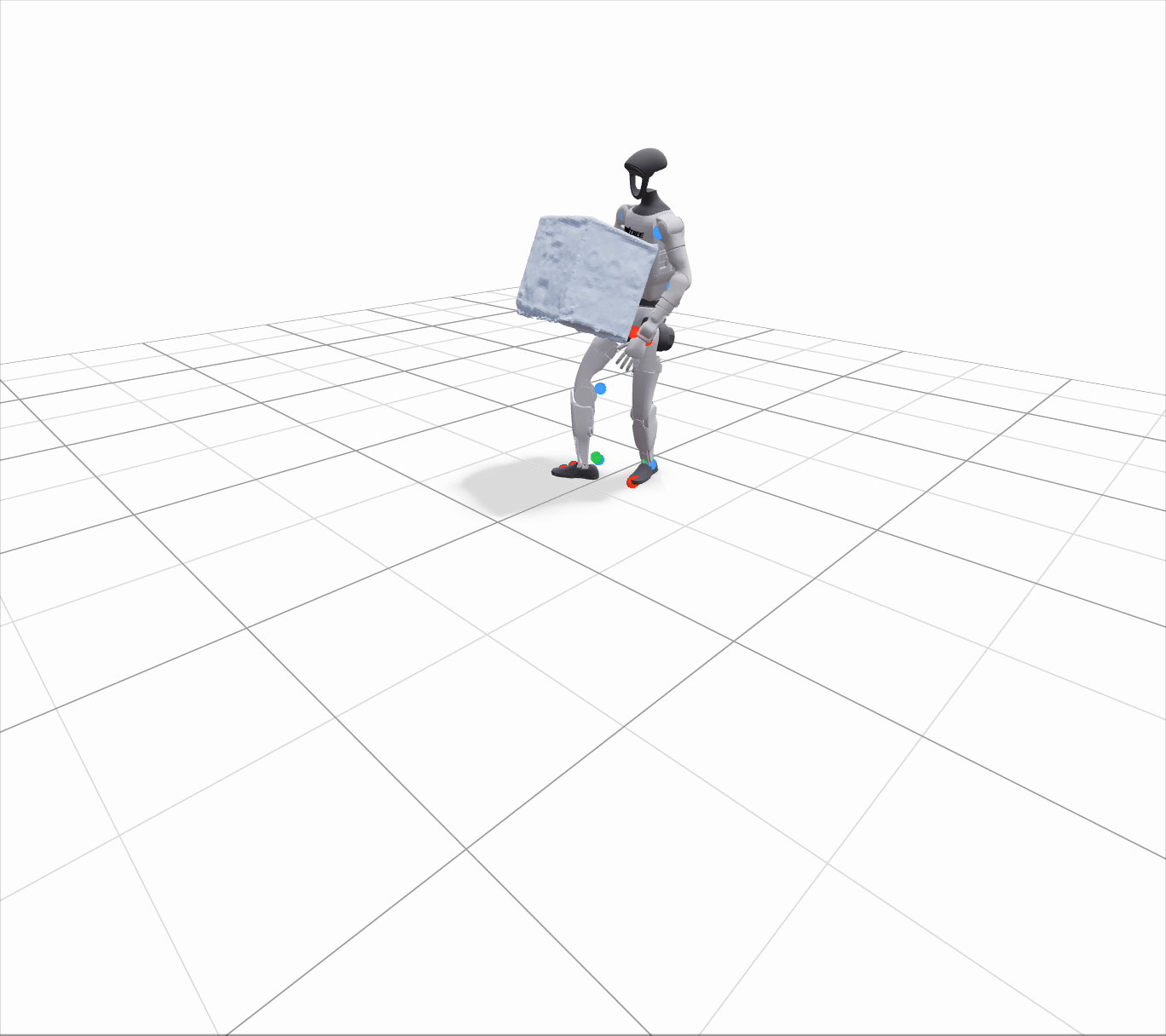}
\captionof{figure}{Retargeted large-box motion at two phases, with body tracking markers visible. The robot reference preserves the interaction with the moving object.}
\label{fig:supp_retarget}
\end{minipage}\par\smallskip

\paragraph{Climbing.}
For climbing demonstrations~\cite{yang2025omniretarget}, we load the sequence-specific terrain geometry and include its surface samples in the interaction mesh. Scene scaling is applied consistently with the motion. Non-penetration and stance-foot constraints preserve support relationships as the robot traverses the terrain, which remains static during tracking.

\paragraph{Sitting.}
For AMASS~\cite{mahmood2019amass}, we first perform robot-only retargeting against the ground. Since the dataset does not provide stool assets, we fit a static stool to each sequence's seated configuration, including its seat height and planar placement. We check seat support, foot placement, and penetration before exporting the fitted furniture and robot motion as a paired reference scene. The stool is a static support surface rather than a moving-object reference.

\section{Egocentric Perception and Deployment}
\label{sec:supp_perception}
\begin{figure*}[!t]
\centering
\includegraphics[width=\textwidth]{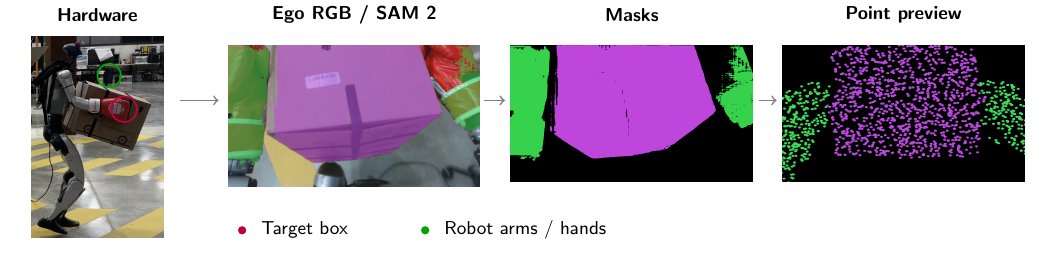}
\caption{\textbf{Real-world egocentric observation processing.} Hardware context, SAM~2 segmentation of ego RGB, semantic masks, and sampled foreground points. Purple denotes the box and green the robot arms/hands. The panels illustrate the processing stages; the point preview is qualitative and shown in image projection.}
\label{fig:supp_real_ego}
\end{figure*}

\paragraph{Real-world observation processing.}
SAM~2~\cite{ravi2024sam2} segments the target object and visible robot arms/hands in ego RGB images using point and box prompts (Figure~\ref{fig:supp_real_ego}). The resulting semantic masks select valid pixels in the aligned depth image. Following the simulation convention, pixel-center back-projection maps a pixel $(u,v)$ with metric depth $d$ to
\begin{equation}
(x,y,z)=\left(d,\frac{c_x-u-0.5}{f_x}d,\frac{c_y-v-0.5}{f_y}d\right),
\end{equation}
where $(f_x,f_y,c_x,c_y)$ are camera intrinsics, and the ego axes point forward, left, and up, respectively. We merge valid foreground points and use farthest-point sampling to obtain 2048 XYZ points, repeating points when fewer are available. Semantic colors are used for visualization, while the policy consumes geometry.

\begin{figure*}[!t]
\centering
\begin{minipage}{.30\textwidth}
\centering
\includegraphics[width=.48\linewidth]{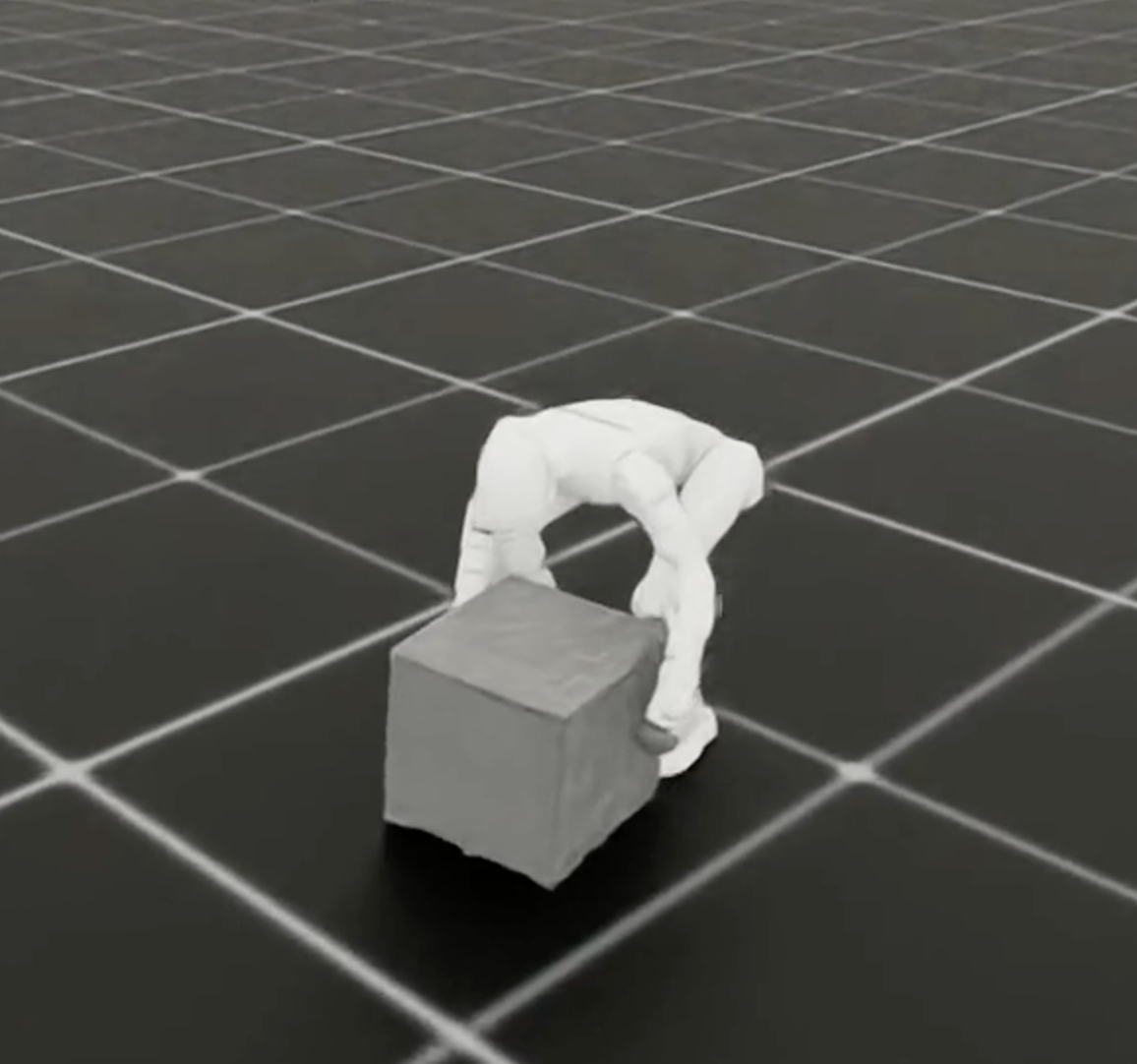}\hfill
\includegraphics[width=.48\linewidth]{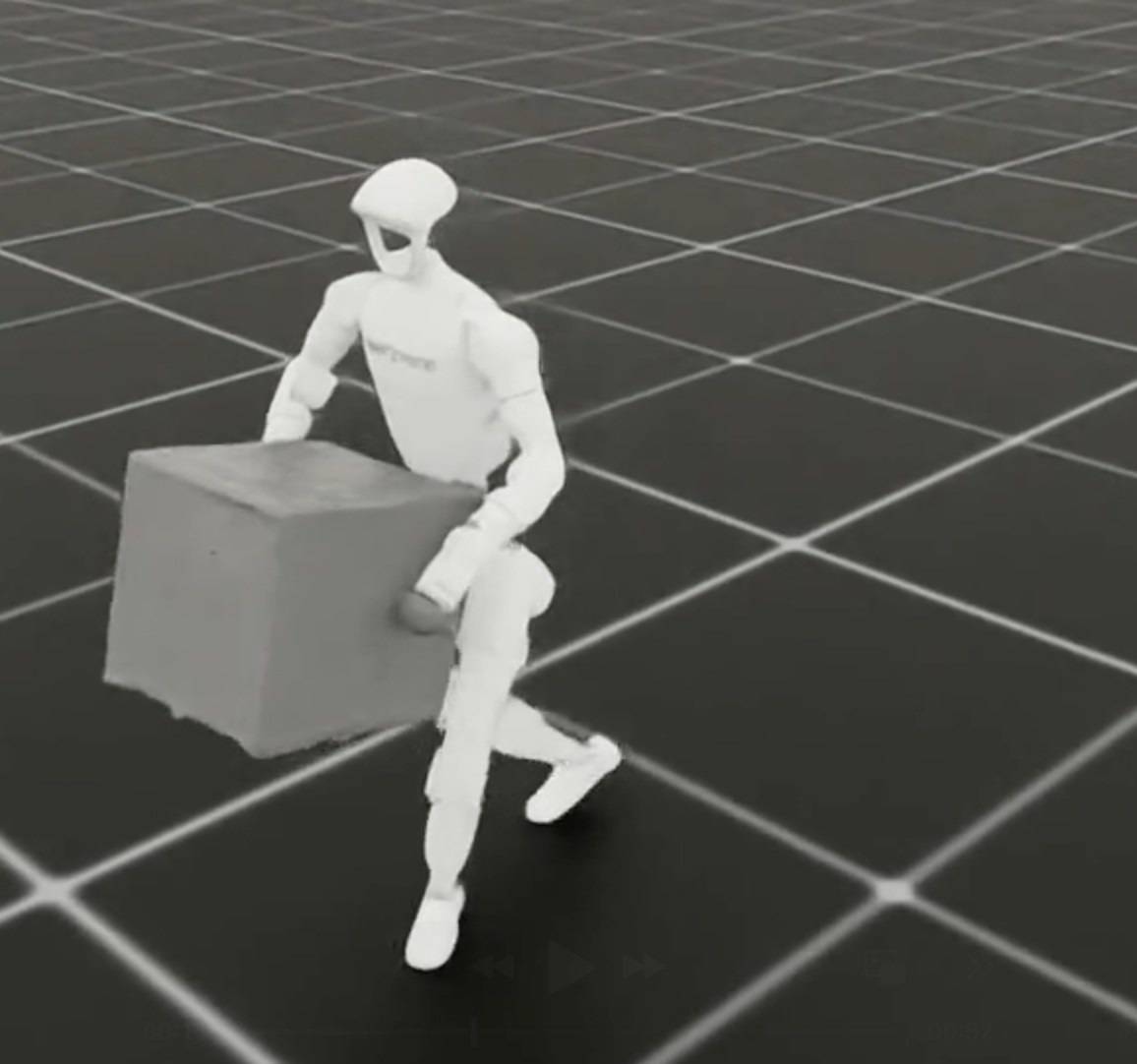}
\end{minipage}\hfill
\begin{minipage}{.68\textwidth}
\includegraphics[width=\linewidth]{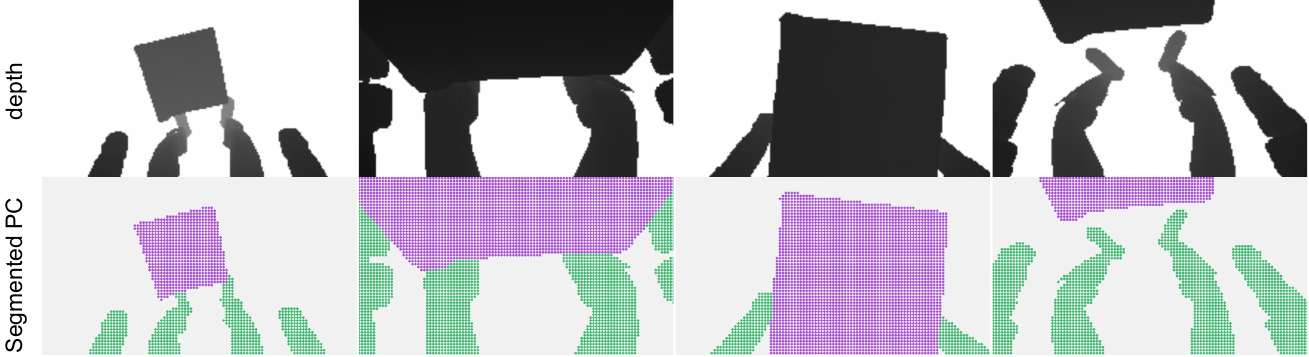}
\end{minipage}
\caption{\textbf{Simulated egocentric observations.} Left: representative box-interaction phases. Right: rendered ego depth (top) and segmented point clouds (bottom) across representative observations. Object and robot-hand points are purple and green. Scene and observation panels illustrate the process rather than synchronized frames.}
\label{fig:supp_ego}
\end{figure*}

\paragraph{Simulation observation processing.}
Rendered depth and simulator segmentation labels provide foreground points in the same ego coordinate convention. For object-interaction distillation, we sample 1024 object points and 1024 hand points. Figure~\ref{fig:supp_ego} illustrates the scene context and the resulting depth and segmented point-cloud observations.

On the G1's Orin NX (16\,GB), TensorRT FP16 with two DDIM steps supports 50\,Hz control and 13.6\,ms median policy latency. Asynchronous perception supplies the latest point cloud alongside robot state.

\section{Additional Real-World Results}
\label{sec:supp_experiments}
\begin{figure*}[!t]
\centering
\includegraphics[width=\textwidth]{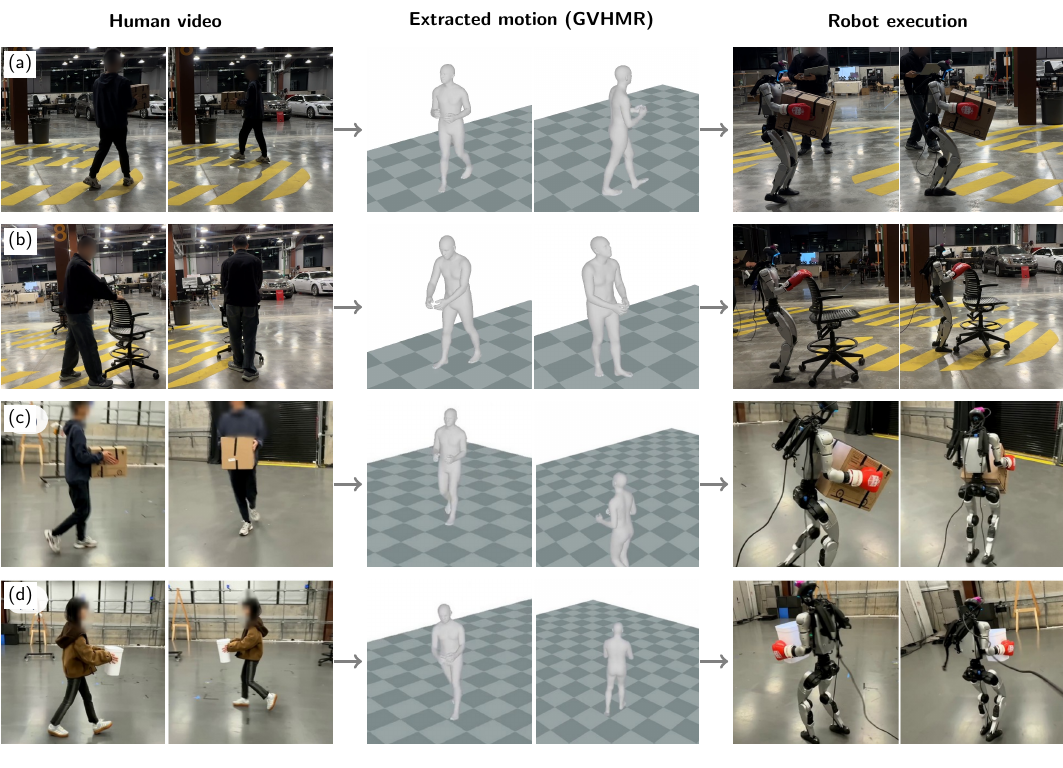}
\caption{\textbf{Video-referenced robot execution.} Each row shows human video frames (left), reconstructed human motion from GVHMR (middle), and robot execution (right). Rows (a,b) show the new box-carrying and chair-pushing examples; rows (c,d) retain the earlier box-carrying and trash-can-carrying examples. Faces are obscured for anonymity.}
\label{fig:supp_real_video}
\end{figure*}

\begin{figure*}[!t]
\centering
\includegraphics[width=\textwidth]{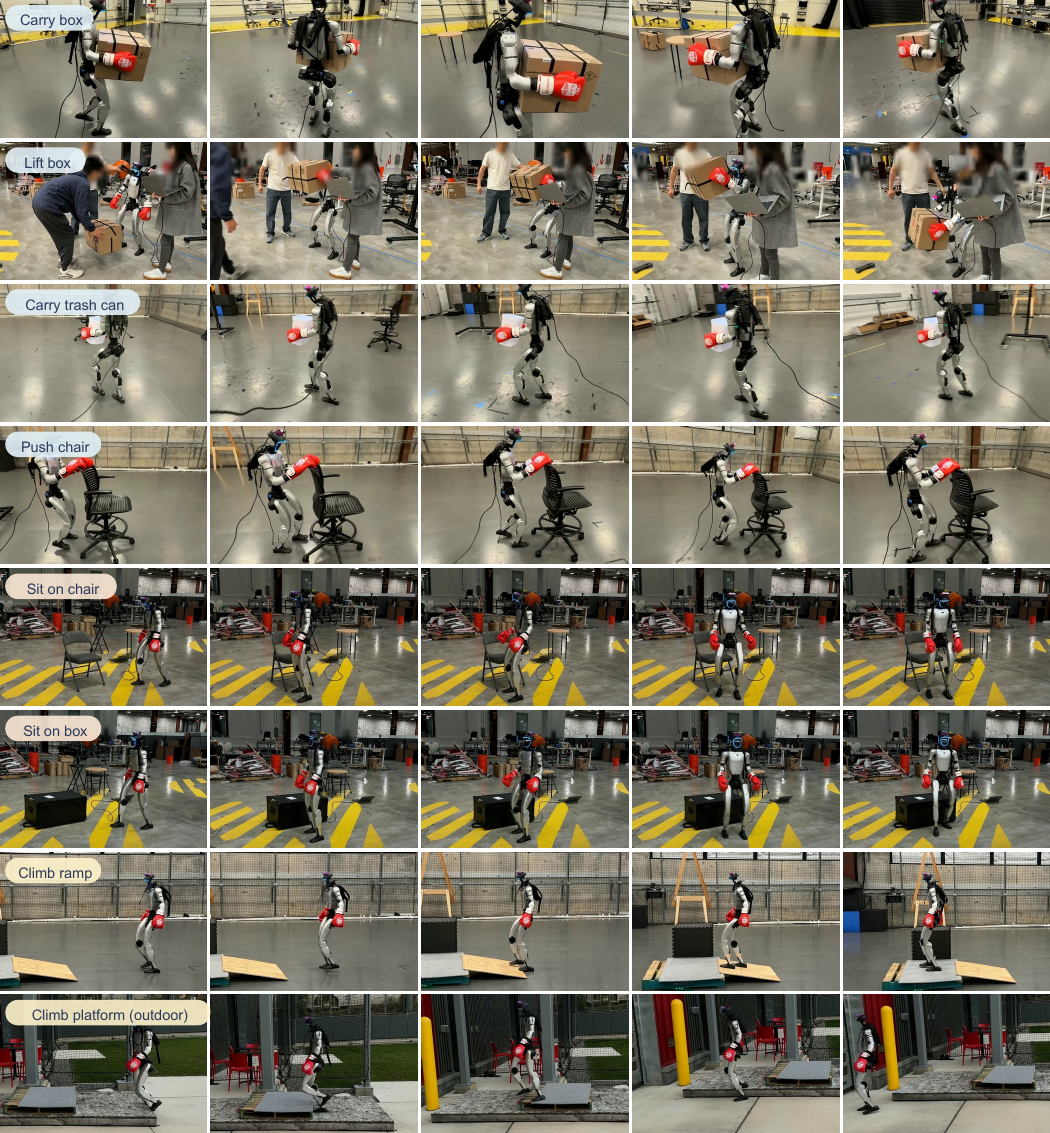}
\caption{\textbf{Additional MoCap-referenced hardware executions.} Rows show box carrying, box lifting, trash-can carrying, chair pushing, sitting on a chair and a box, and climbing a ramp and an outdoor platform. Frames progress left to right within each row. All executions use the same frozen \projname{} policy.}
\label{fig:supp_real_all}
\end{figure*}

The same frozen \projname{} policy transfers from simulation to the Unitree G1 with both MoCap and video-derived references. Figure~\ref{fig:supp_real_video} shows new box-carrying and chair-pushing demonstrations alongside the earlier box-carrying and trash-can-carrying examples, pairing human videos, GVHMR reconstructions, and robot executions. The reconstructed human motions provide inputs for robot retargeting. Figure~\ref{fig:supp_real_all} extends the MoCap-referenced examples to moving-object interactions and whole-body support transitions, including different seating surfaces and indoor/outdoor terrain. These sequences illustrate the range of interactions supported by one policy without task-specific fine-tuning. Full-length executions are provided in the supplementary video.

\section{Additional Simulation Results}
\label{sec:supp_sim}
\begin{figure*}[!t]
\centering
\includegraphics[width=\textwidth]{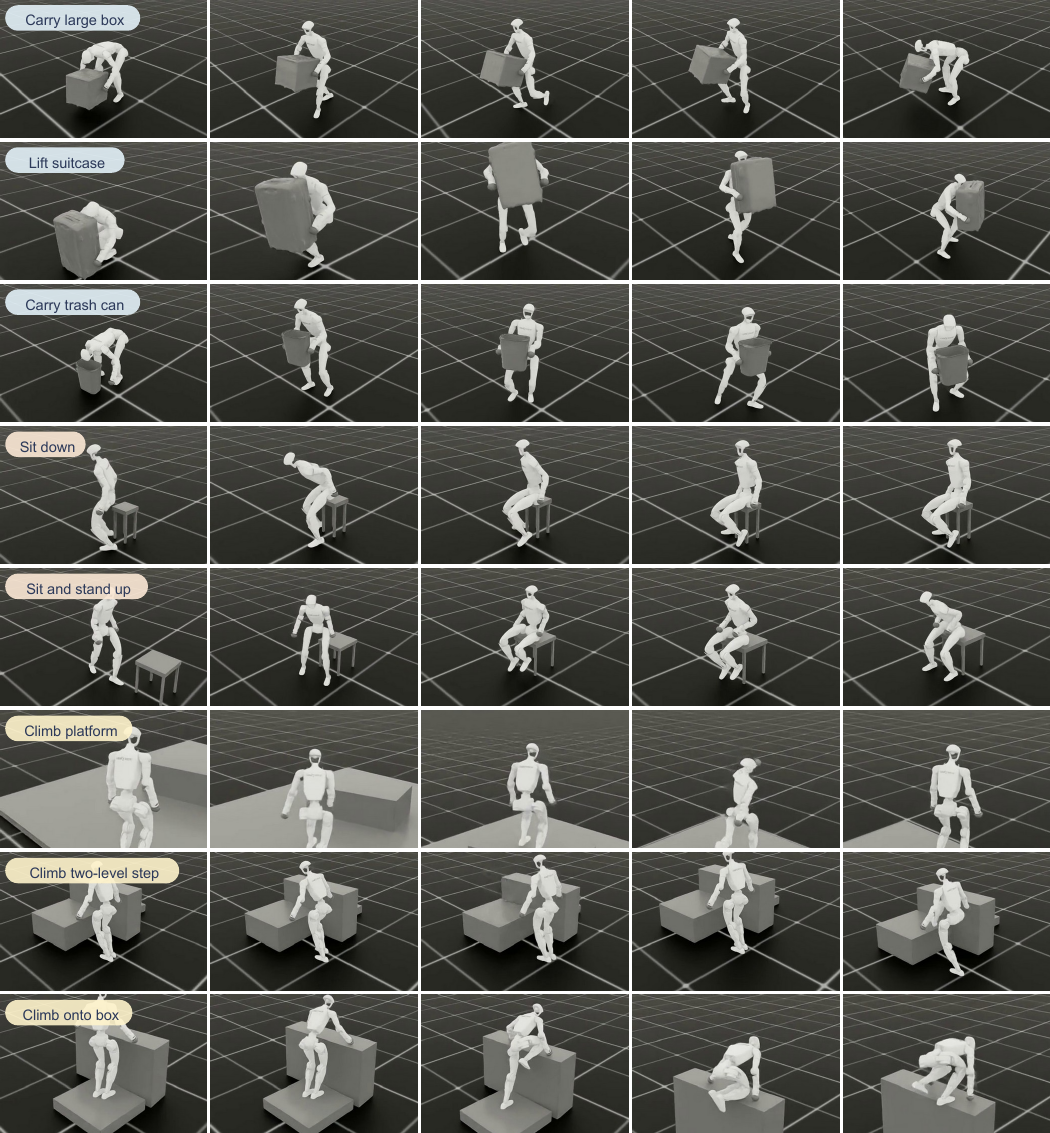}
\caption{\textbf{Simulation executions on held-out references.} Rows show large-box carrying, suitcase lifting, trash-can carrying, two sitting sequences, and climbing on a platform, a two-level step, and a box. Each row samples one rollout from left to right. All use the same distilled \projname{} policy.}
\label{fig:supp_sim_all}
\end{figure*}

Figure~\ref{fig:supp_sim_all} presents additional distilled-student rollouts on held-out references, spanning moving-object manipulation and changes in whole-body support during sitting and climbing. The sequences illustrate coordinated arm, torso, and leg motion under a shared reference-conditioned policy. Figure~\ref{fig:supp_sim_parallel} additionally shows parallel evaluation across climbing and sitting references: each environment follows its own reference while sharing the same student parameters.

\clearpage
\twocolumn[{\begin{minipage}{\textwidth}
\captionsetup{hypcap=false}
\centering
\includegraphics[width=\textwidth]{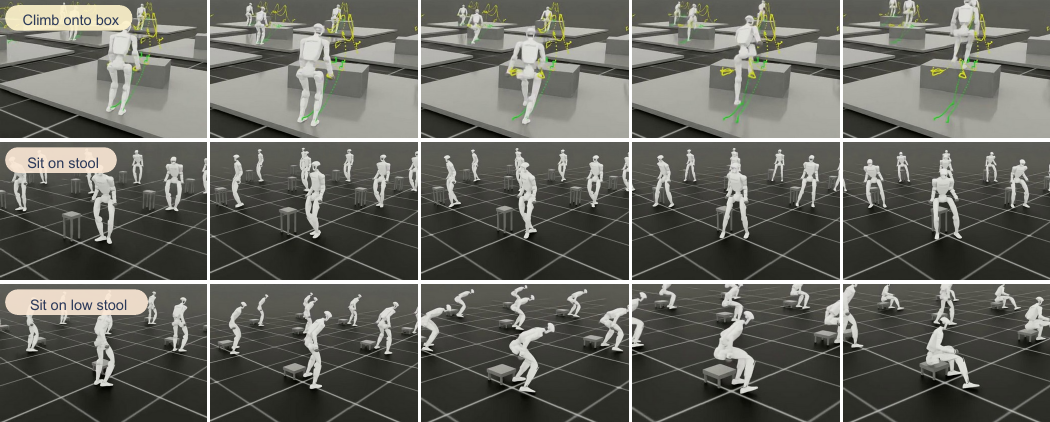}
\captionof{figure}{\textbf{Shared-policy evaluation in parallel environments.} Rows show climbing onto a box and sitting on stools of different heights, sampled left to right. Multiple robots execute their respective references simultaneously using the same distilled student policy.}
\label{fig:supp_sim_parallel}
\end{minipage}\par\vspace{10pt}}]
\section{Discussion: Scalable Tracking with SuperMimic vs. Grouped RL Experts}
\label{sec:supp_supermimic}
Directly training a world-model-based generalist with SuperMimic~\cite{liao2026supermimic} does not efficiently master the full evaluated interaction set. Across nine OMOMO objects, mean success remains around 0.5 after 686M transitions (72 GPU-hours), while long-tail objects remain at 0--0.5 (Table~\ref{tab:supp_supermimic}). This motivates learning interaction-specific experts before visuomotor distillation.

\paragraph{Motion tracking versus interaction.}
Despite SuperMimic's strong reported performance on robot-only motion tracking~\cite{liao2026supermimic}, reproducing body motion alone does not ensure successful object interaction. In our physical-box simulation (Figure~\ref{fig:supp_supermimic_box}), the policy continues the reference body motion after losing the box, failing to maintain the intended carrying interaction. This example highlights the need to learn contact-consistent robot--object dynamics alongside motion tracking.

\par\smallskip
\noindent\begin{minipage}{\linewidth}
\captionsetup{hypcap=false}
\centering
\includegraphics[height=2.25cm]{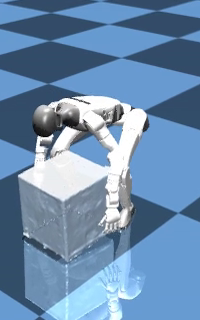}\hfill
\includegraphics[height=2.25cm]{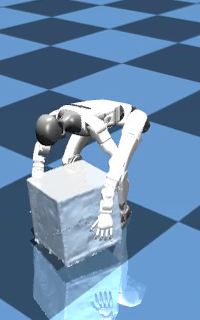}\hfill
\includegraphics[height=2.25cm]{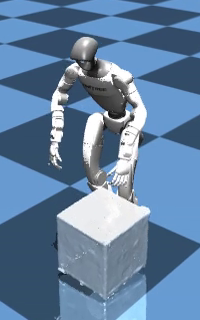}\hfill
\includegraphics[height=2.25cm]{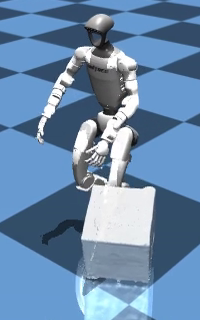}\hfill
\includegraphics[height=2.25cm]{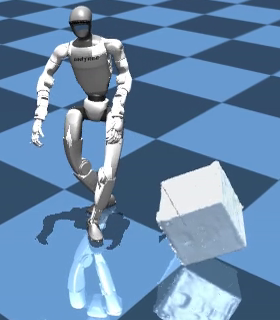}
\captionof{figure}{\textbf{Loss of object interaction.} SuperMimic continues the body motion as the box slips and falls, shown left to right.}
\label{fig:supp_supermimic_box}
\end{minipage}\par\smallskip

\paragraph{Success criterion.}
We use the same success criterion as the main-paper benchmark. A clip succeeds only if none of the following failure thresholds is exceeded throughout its execution: torso height error, 0.5\,m; torso tilt error $|g_z-\bar g_z|$, 0.8; any wrist or ankle height error, 0.25\,m; object position error, 0.25\,m; or object orientation error, 0.8\,rad. Here $g_z$ and $\bar g_z$ are the gravity-direction projections onto the actual and reference torso $z$ axes. This evaluates full-trajectory success, rather than partial trajectory completion.

\par\smallskip
\noindent\begin{minipage}{\linewidth}
\captionsetup{hypcap=false}
\centering\small
\captionof{table}{Direct generalist RL on nine OMOMO objects. Simple objects: box, table, trash can; long-tail objects: chairs, monitor, suitcase. Success rates are fractions.}
\label{tab:supp_supermimic}
\setlength{\tabcolsep}{3pt}
\begin{tabularx}{\linewidth}{@{}Xccc@{}}
\toprule
Training (steps / GPU-h) & Mean & Simple & Long-tail \\
\midrule
205M / 16 & 0.28 & 0.6--0.9 & 0--0.3 \\
410M / 33 & 0.47 & 0.9--1.0 & 0--0.2 \\
686M / 72 & 0.44--0.54 & 0.9--1.0 & 0--0.5 \\
\bottomrule
\end{tabularx}
\end{minipage}\par\smallskip

\end{document}